\documentclass[10pt,twocolumn,letterpaper]{article}
\usepackage{graphicx}
\usepackage{subcaption}
\usepackage[svgnames]{xcolor}
\usepackage{tikz}
\usepackage{booktabs}
\usepackage[margin=1in]{geometry}
\usetikzlibrary{arrows.meta,positioning,fit,shadows.blur}
\usepackage{graphicx}
\usepackage{xcolor}
\usepackage{booktabs}
\usepackage{caption}

\usepackage{dblfloatfix}

\usepackage{graphicx}
\usepackage{amsmath}
\usepackage{amssymb}
\usepackage{booktabs}
\usepackage{tikz}
\usetikzlibrary{arrows.meta,positioning,fit,shadows.blur}
\usepackage{graphicx}
\usepackage{xcolor}
\usepackage{booktabs}
\usepackage{caption}
\usepackage[toc,page]{appendix}

\usepackage[most]{tcolorbox}  % main package for colored/shaded boxes
\usepackage{xcolor}           % extended color support
\usepackage{graphicx}         % (optional) if you later include example images
\usepackage{multirow}
\usepackage{adjustbox} % Add this line
\usepackage{multirow} % Add this line at the beginning of your LaTe
\usepackage{makecell}
\usepackage{pifont}

\usepackage[utf8]{inputenc} % allow utf-8 input
\usepackage[T1]{fontenc}    % use 8-bit T1 fonts
\usepackage{booktabs}       % professional-quality tables
\usepackage{amsfonts}       % blackboard math symbols
\usepackage{nicefrac}       % compact symbols for 1/2, etc.
\usepackage{microtype}      % microtypography
\usepackage{xcolor}         % colors
\usepackage{graphicx} % for \includegraphics
\usepackage{subcaption} % for subfigure environment
\usepackage{booktabs} % for \toprule, \midrule, \bottomrule
\usepackage{caption} % for \captionsetup

\usepackage{pifont}
\usepackage[dvipsnames]{xcolor}
\usepackage{colortbl}

\usepackage{graphicx}
\usepackage{booktabs}
\usepackage{tcolorbox}
\usepackage[most]{tcolorbox}

\definecolor{cvprblue}{rgb}{0.21,0.49,0.74}

\usepackage{cvpr}              % To produce the CAMERA-READY version
\usepackage{tikz}
\usetikzlibrary{arrows.meta,positioning,fit,shadows.blur}
\usepackage{graphicx}
\usepackage{xcolor}
\usepackage{booktabs}
\usepackage{caption}

\definecolor{cvprblue}{rgb}{0.21,0.49,0.74}
\usepackage[pagebackref,breaklinks,colorlinks,allcolors=cvprblue]{hyperref}

\def\paperID{} % *** Enter the Paper ID here
\def\confName{CVPR}
\def\confYear{2026}

\newcommand{\ours}{FFGo}

\title{
First Frame Is the Place to Go for Video Content Customization}

\author{
Jingxi Chen$^{1*\dagger}$, Zongxia Li$^{1*\dagger}$, Zhichao Liu, Guangyao Shi$^{2}$, Xiyang Wu$^{1}$, Fuxiao Liu$^{4}$, \\  
Cornelia Ferm{\"u}ller$^{1}$, Brandon Y. Feng$^{3\dagger}$, Yiannis Aloimonos$^{1}$ \\[3pt]
$^{1}$University of Maryland \quad $^{2}$USC \quad $^{3}$MIT \quad 
$^{4}$NVIDIA \\[2pt]
\url{http://firstframego.github.io}
}

\begin{document}

\twocolumn[{%
\renewcommand\twocolumn[1][]{#1}%
\maketitle

\vspace{-28pt}

\begin{center}
    \includegraphics[scale=0.34]{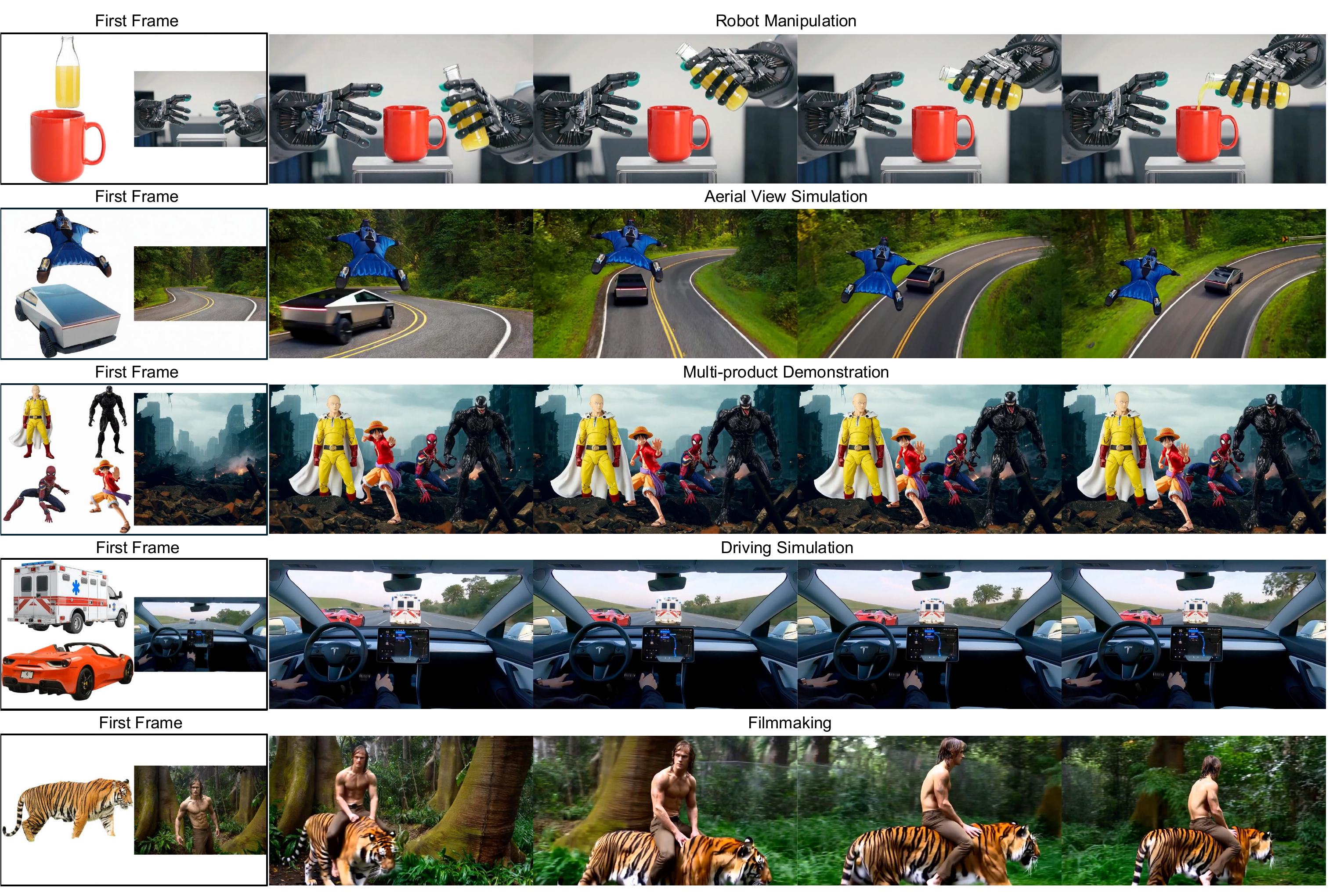} % Adjust the scale as needed
    \vspace{-25pt}
    \captionsetup{type=figure}
    \caption{\ours{} is a lightweight add-on that invokes the innate capabilities of pre-trained video generation models, such as Wan2.2 \cite{wan2025wanopenadvancedlargescale}, to treat the first frame as a compositional blueprint, enabling natural subject mixing and interaction throughout the video. Given a single input image with multiple elements and a guiding text prompt, FFGo generates coherent, customized videos across diverse applications including robotic manipulation, driving/aerial/underwater simulation, multi-product demonstration, and filmmaking. It requires no architectural modifications, and achieves strong subject-mixing performance with just 20–50 LoRA fine-tuning examples.}
    \vspace{-5pt}
    \captionsetup{font=small}
    \label{fig:teaser}
    % \vspace{-0pt}
\end{center}%
}]

\begin{abstract}

\begingroup
\renewcommand\thefootnote{}\footnote{*Equal contributions $\dagger$ Corresponding author.}
\addtocounter{footnote}{-1}
\endgroup
\vspace{-18pt}

What role does the first frame play in video generation models? Traditionally, it’s viewed as the spatial-temporal starting point of a video, merely a seed for subsequent animation. In this work, we reveal a fundamentally different perspective: video models implicitly treat the first frame as a conceptual memory buffer that stores visual entities for later reuse during generation. Leveraging this insight, we show that it's possible to achieve robust and generalized video content customization in diverse scenarios, using only 20–50 training examples without architectural changes or large-scale finetuning. This unveils a powerful, overlooked capability of video generation models for reference-based video customization.

\end{abstract}

\section{Introduction}
\label{sec:intro}

Recent advances in video generation models~\cite{blattmann2023stable, polyak2024movie, yang2024cogvideox, liu2024sora, multimodalVideoGen, wan2025wanopenadvancedlargescale, wiedemer2025videomodelszeroshotlearners, burgert2025go, chen2025repurposing, yuan2025identity, zhang2024physdreamer} have made them powerful tools for content creation, filmmaking, simulation, and other applications involving creative visual experiences.
A key application of these models is reference-based video generation, where one or more reference inputs are used to compose and synthesize visually consistent videos. This capability is essential for real-world scenarios such as film production and simulation, where customization and controllability are critical. Unlike standard Text-to-Video (T2V) models that rely solely on textual prompts, reference-based generation allows users to guide video synthesis through visual references, enabling finer control over generated video contents to be customized for the specific user guidance.

The simplest case is using a single reference image, known as the Image-to-Video (I2V) paradigm. I2V models animate a given image to generate videos that maintain visual consistency with the reference content. While effective as a basic form of video customization, this single-image setup constrains both the spatial diversity and content composition of the generated video. To overcome these limitations, recent research has focused on developing multi-reference video generation models~\cite{jiang2025vaceallinonevideocreation, fei2025skyreelsa2composevideodiffusion, liao2025character, chen2025multi, wang2025dreamactor}, which can incorporate multiple reference images to achieve richer and more flexible video content customization.

Existing multi-reference video generation models generally follow two strategies: 1) Architectural modification of pre-trained video generation models to accommodate additional reference inputs. 2) Fine-tuning on large-scale, task-specific video customization datasets, such as inserting humans into animations or product demonstration videos. However, fine-tuning often leads to performance degradation and loss of generalization in the adapted models. Since video customization typically occurs post-training, the diversity and quality of the adaptation datasets are far lower than those used during pre-training. Consequently, fine-tuned models tend to overfit to specific customization scenarios, forgetting the broad generative priors learned during pre-training, effectively transforming once generalist video models into narrow, task-specialized systems.

\textit{
Is it possible to incorporate content from multiple reference images into pre-trained video generation models without modifying their architecture or relying on large-scale video training datasets?}

\begin{figure*}[htbp]
    \centering
    \begin{tikzpicture}
        \node[fill=LightGrey!40, inner sep=2pt, text width=0.98\linewidth] {
            \parbox{1\linewidth}{
                \centering
                \makebox[0.19\linewidth][c]{\small\textbf{First Frame}}\hspace{0.9em}%
                \makebox[0.78\linewidth][c]{\small\textbf{Veo 3}} \\[2pt]
                % Wrap first frame in tikz node for alignment
                \tikz[baseline={(current bounding box.center)}]{
                    \node[inner sep=0pt] {
                        \includegraphics[width=0.18\linewidth]{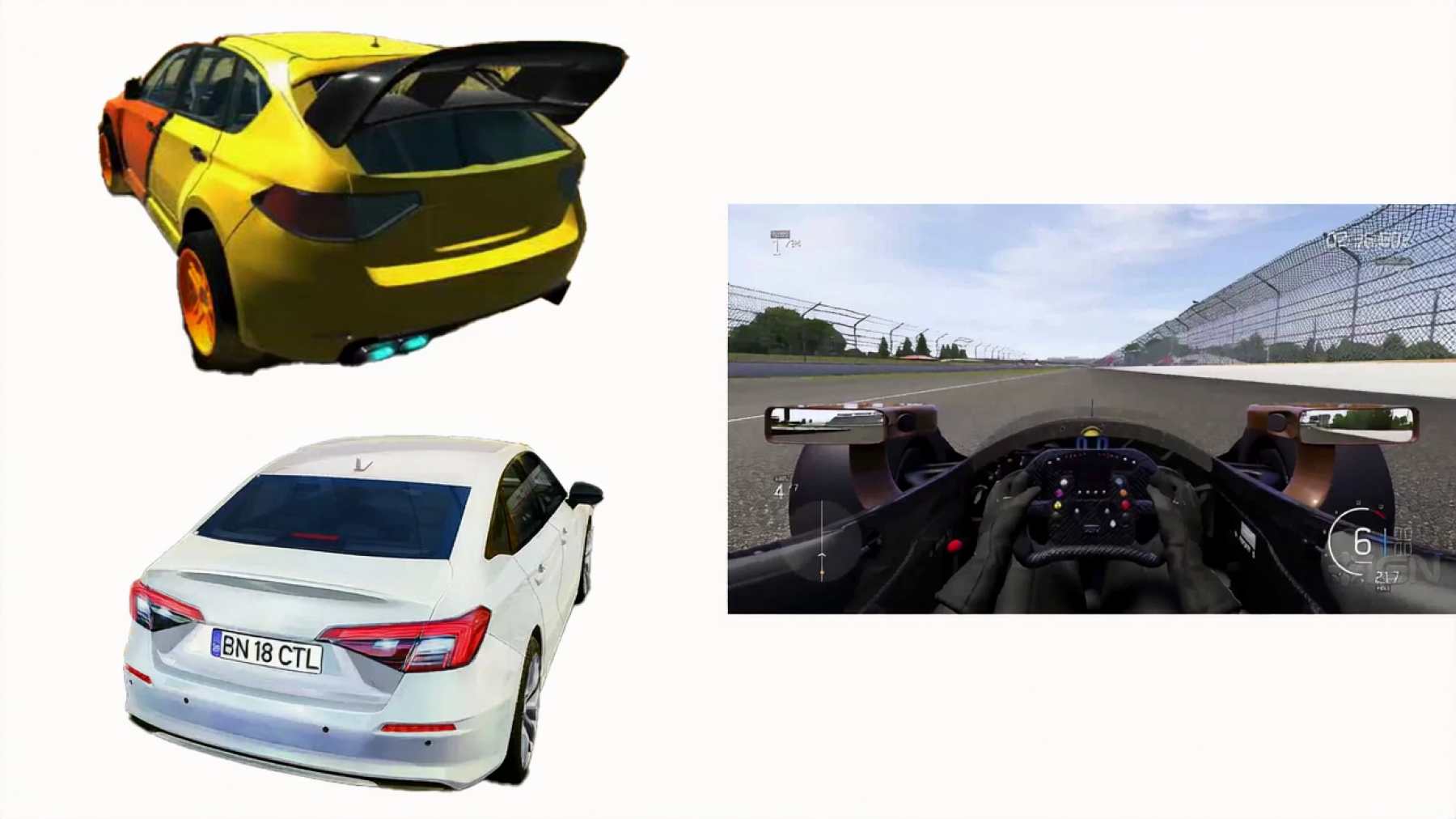}
                    };
                }%
                \hspace{0.9em}%
                % Rectangle around frames 2-3
                \tikz[baseline={(current bounding box.center)}]{
                    \node[draw=red, line width=2pt, inner sep=2pt] {
                        \includegraphics[width=0.18\linewidth]{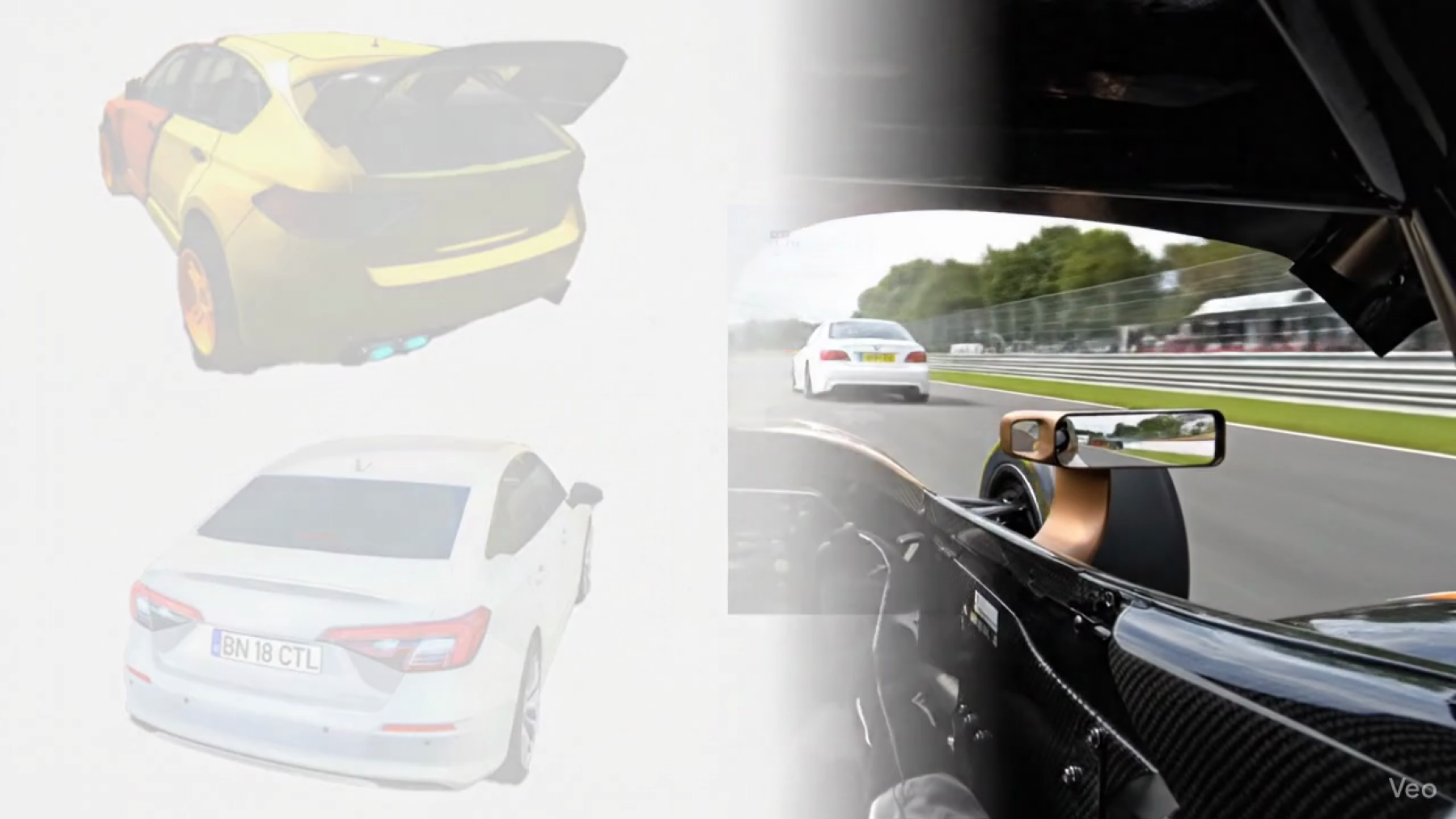}\hspace{0.3em}%
                        \includegraphics[width=0.18\linewidth]{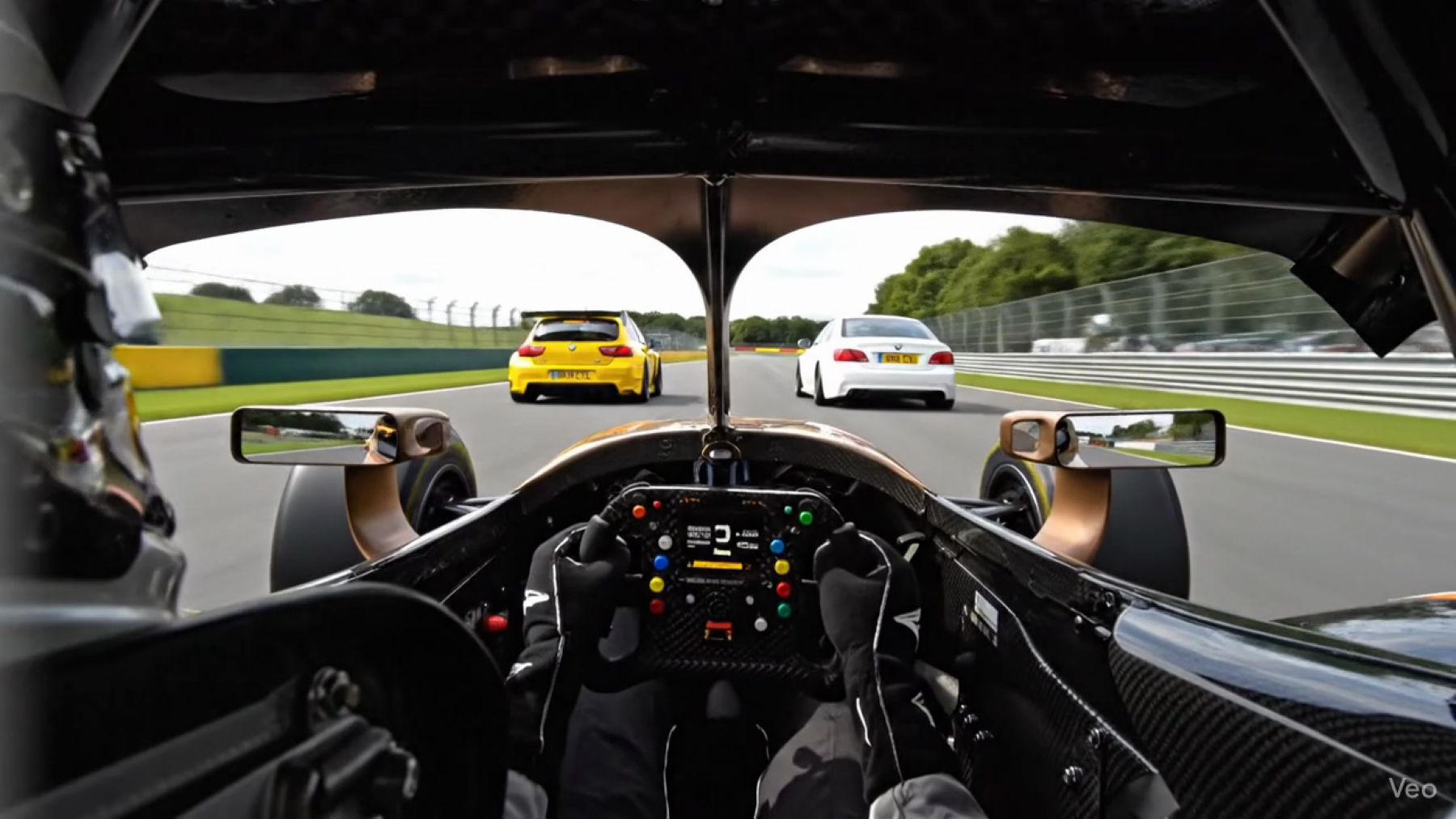}
                    };
                }%
                \hspace{0.3em}%
                % Rectangle around frames 4-5
                \tikz[baseline={(current bounding box.center)}]{
                    \node[draw=blue, line width=2pt, inner sep=2pt] {
                        \includegraphics[width=0.18\linewidth]{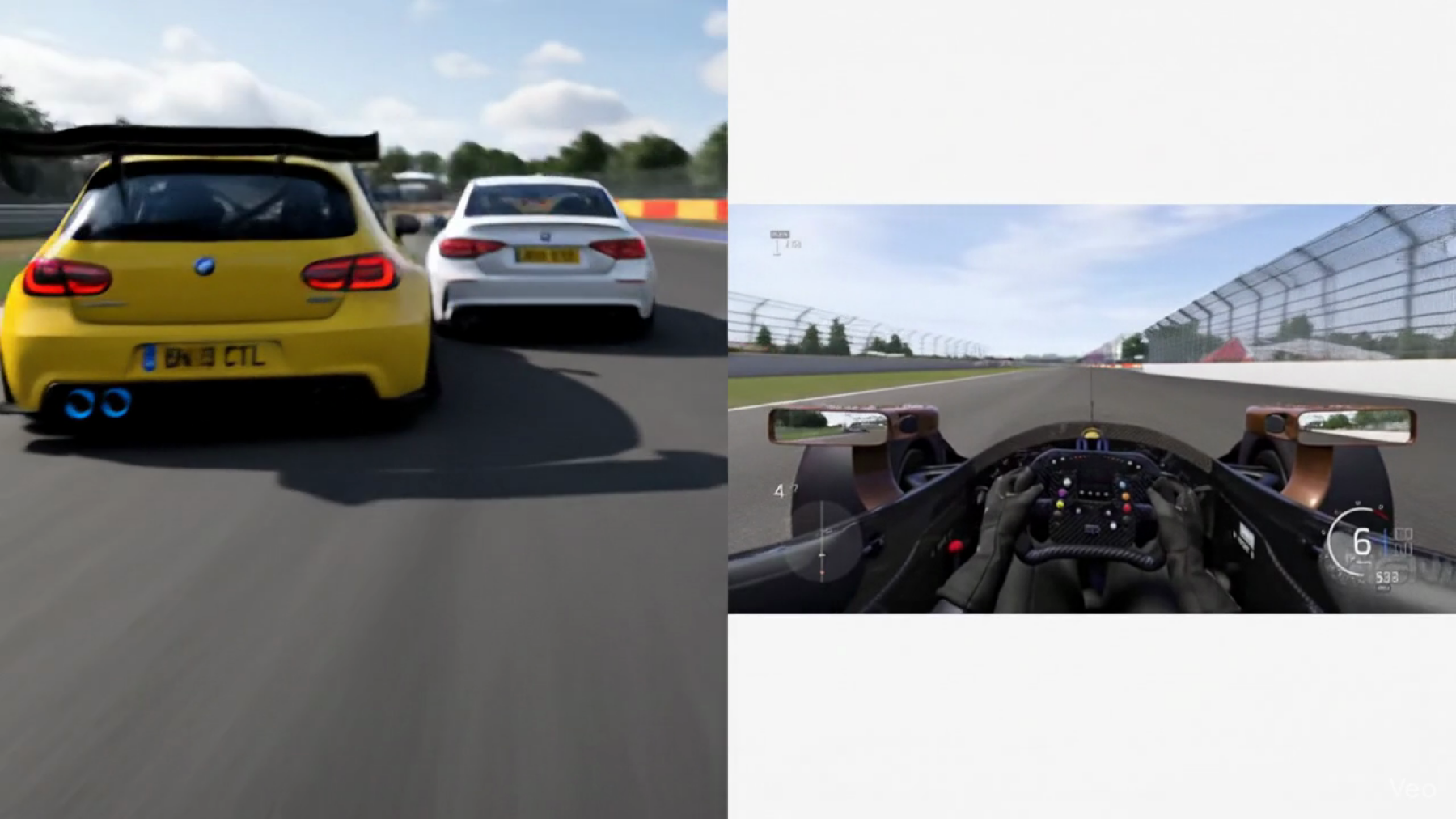}\hspace{0.3em}%
                        \includegraphics[width=0.18\linewidth]{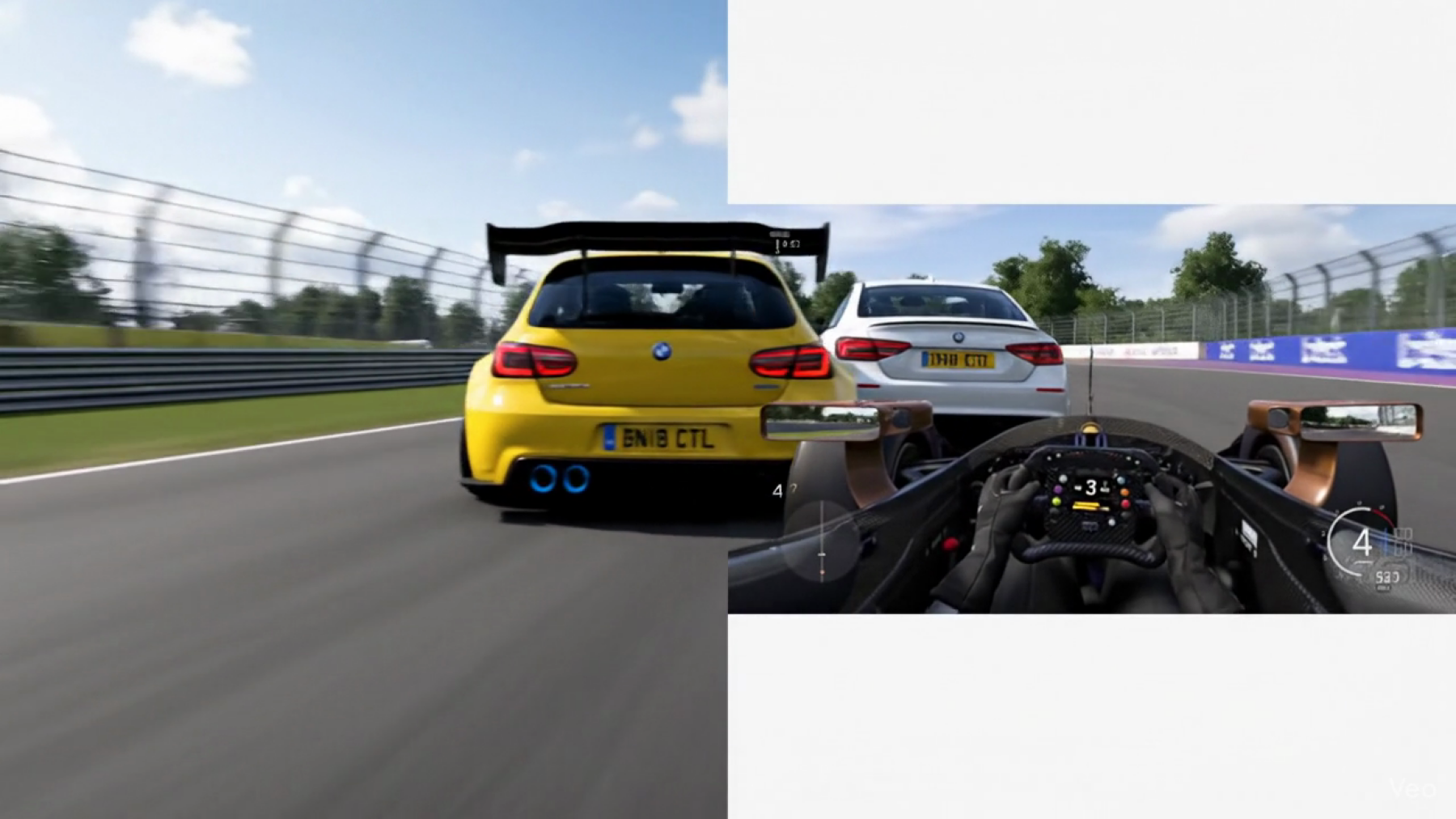}
                    };
                }%
            }
        };
    \end{tikzpicture}
    
    % Second Row - Aerial View Simulation
    \begin{tikzpicture}
        \node[fill=LightGrey!40, inner sep=2pt, text width=0.98\linewidth] {
            \parbox{1\linewidth}{
                \centering
                \makebox[0.19\linewidth][c]{\small\textbf{First Frame}}\hspace{0.9em}%
                \makebox[0.78\linewidth][c]{\small\textbf{Sora 2}} \\[2pt]
                \tikz[baseline={(current bounding box.center)}]{
                    \node[inner sep=0pt] {
                        \includegraphics[width=0.18\linewidth]{figures/transitions/wan_transition_frame_0000.png}
                    };
                }%
                \hspace{0.9em}%
                \tikz[baseline={(current bounding box.center)}]{
                    \node[draw=red, line width=2pt, inner sep=2pt] {
                        \includegraphics[width=0.18\linewidth]{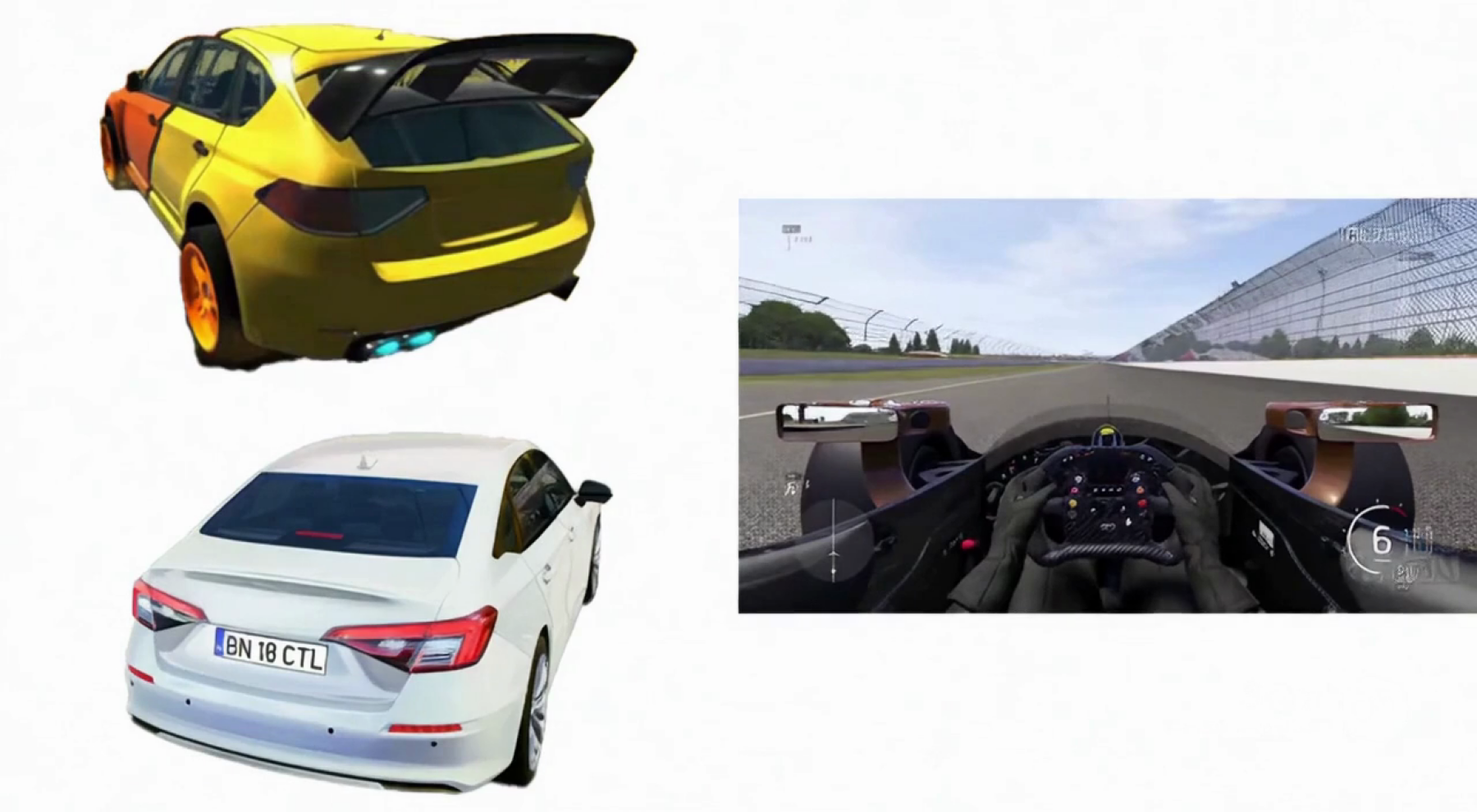}\hspace{0.3em}%
                        \includegraphics[width=0.18\linewidth]{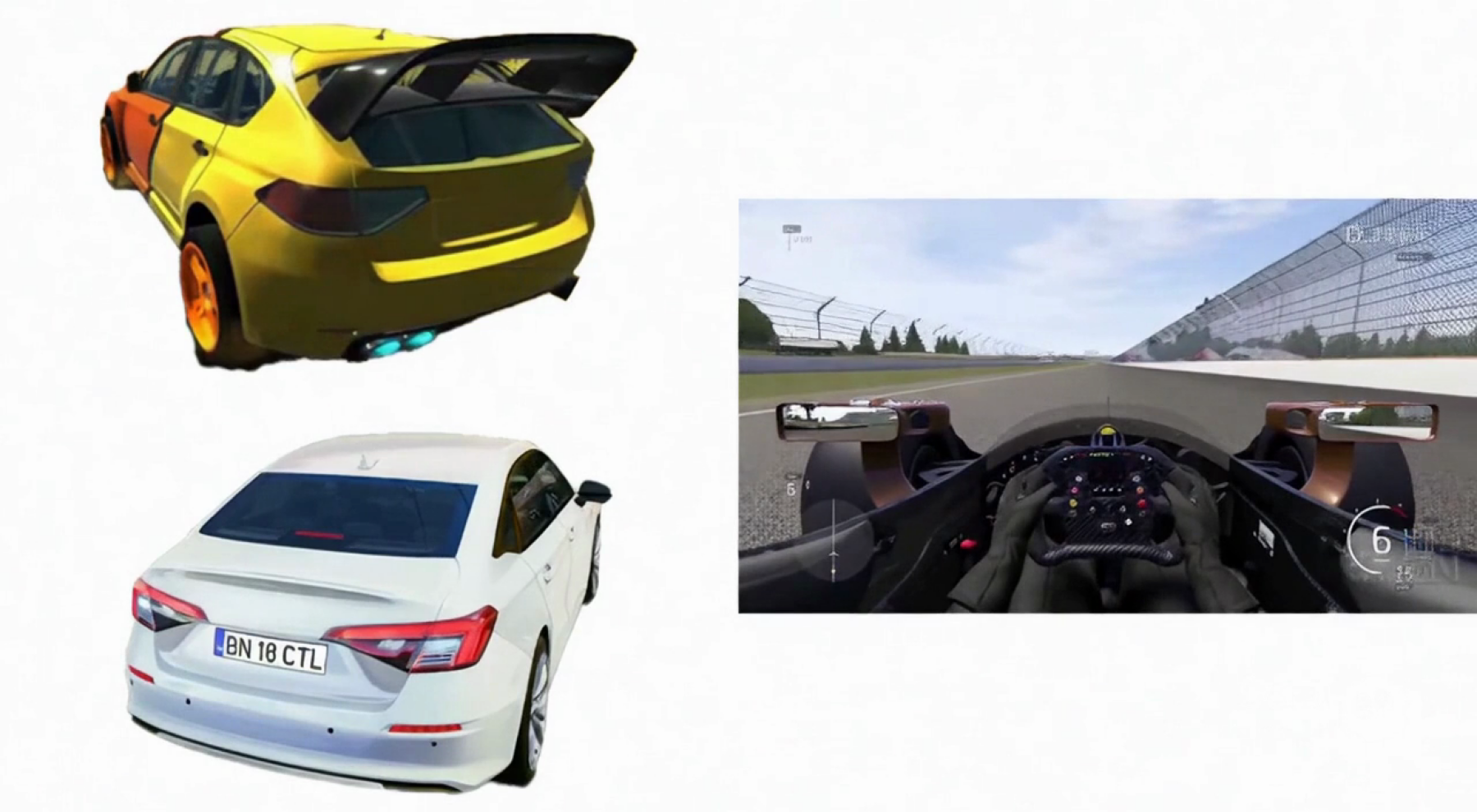}
                    };
                }%
                \hspace{0.3em}%
                \tikz[baseline={(current bounding box.center)}]{
                    \node[draw=blue, line width=2pt, inner sep=2pt] {
                        \includegraphics[width=0.18\linewidth]{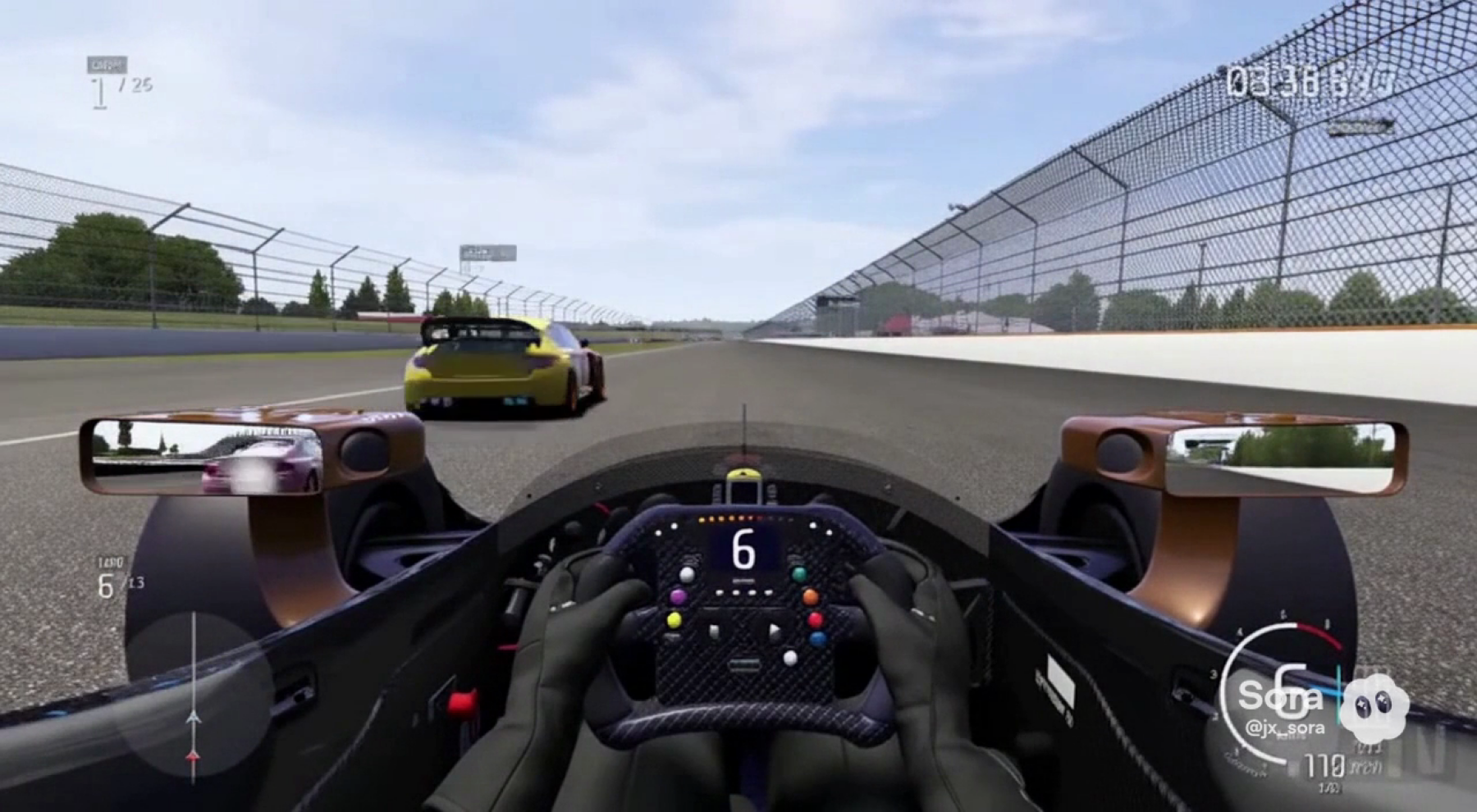}\hspace{0.3em}%
                        \includegraphics[width=0.18\linewidth]{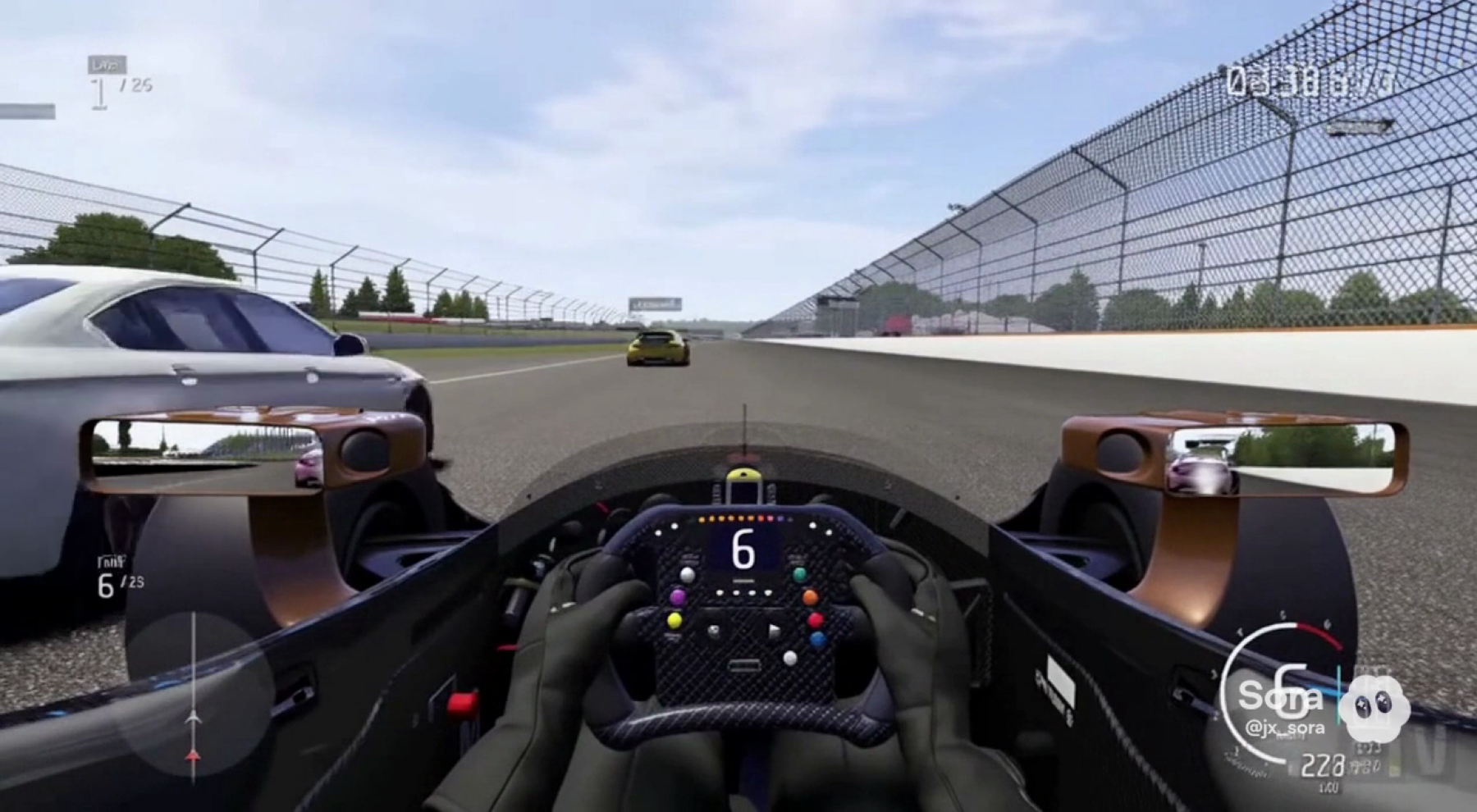}
                    };
                }%
            }
        };
    \end{tikzpicture}
    
    % Third Row - Multi-object Mixing
    \begin{tikzpicture}
        \node[fill=LightGrey!40, inner sep=2pt, text width=0.98\linewidth] {
            \parbox{1\linewidth}{
                \centering
                \makebox[0.19\linewidth][c]{\small\textbf{First Frame}}\hspace{0.9em}%
                \makebox[0.78\linewidth][c]{\small\textbf{Wan 2.2}} \\[2pt]
                \tikz[baseline={(current bounding box.center)}]{
                    \node[inner sep=0pt] {
                        \includegraphics[width=0.18\linewidth]{figures/transitions/wan_transition_frame_0000.png}
                    };
                }%
                \hspace{0.9em}%
                \tikz[baseline={(current bounding box.center)}]{
                    \node[draw=red, line width=2pt, inner sep=2pt] {
                        \includegraphics[width=0.18\linewidth]{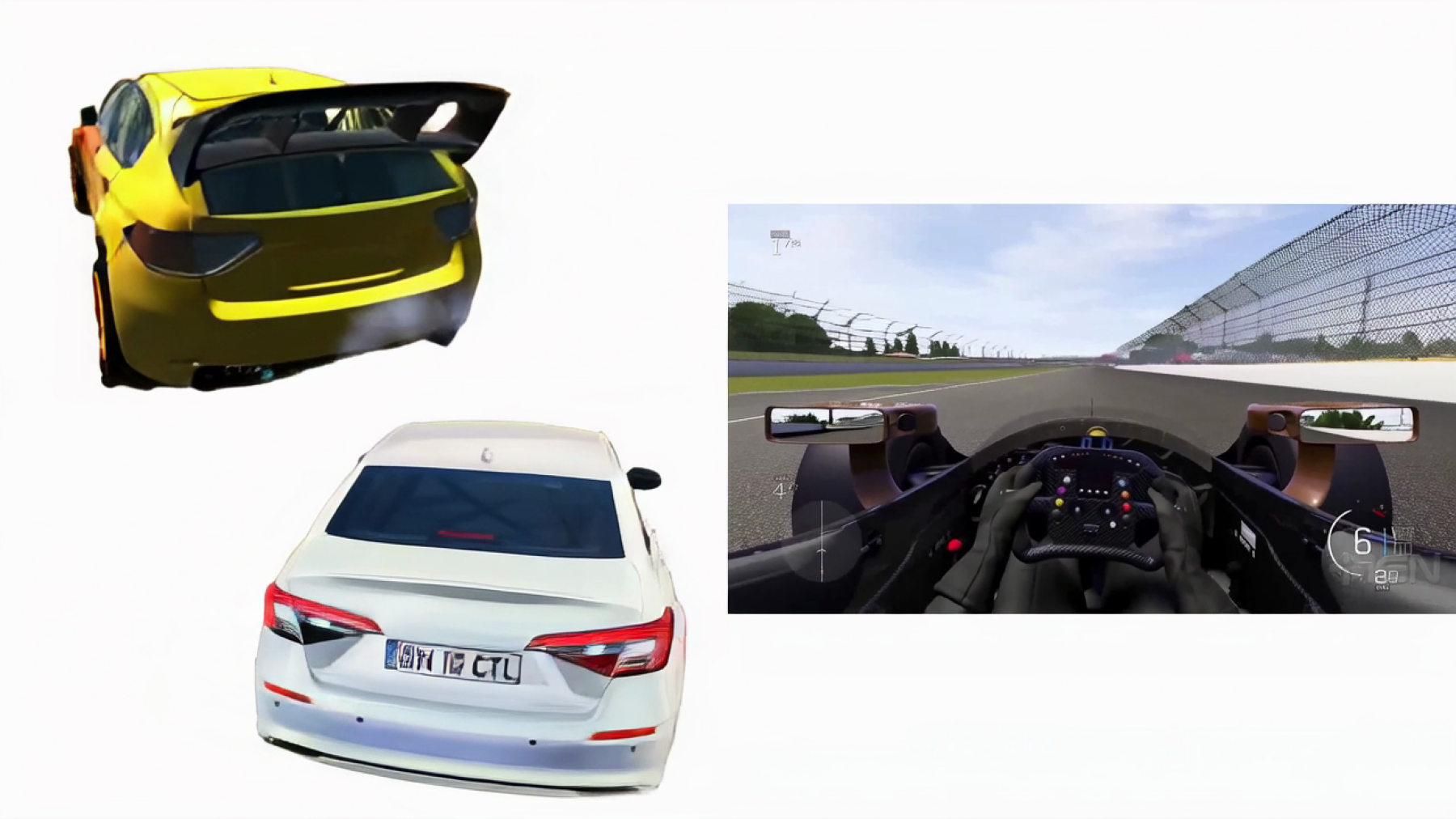}\hspace{0.3em}%
                        \includegraphics[width=0.18\linewidth]{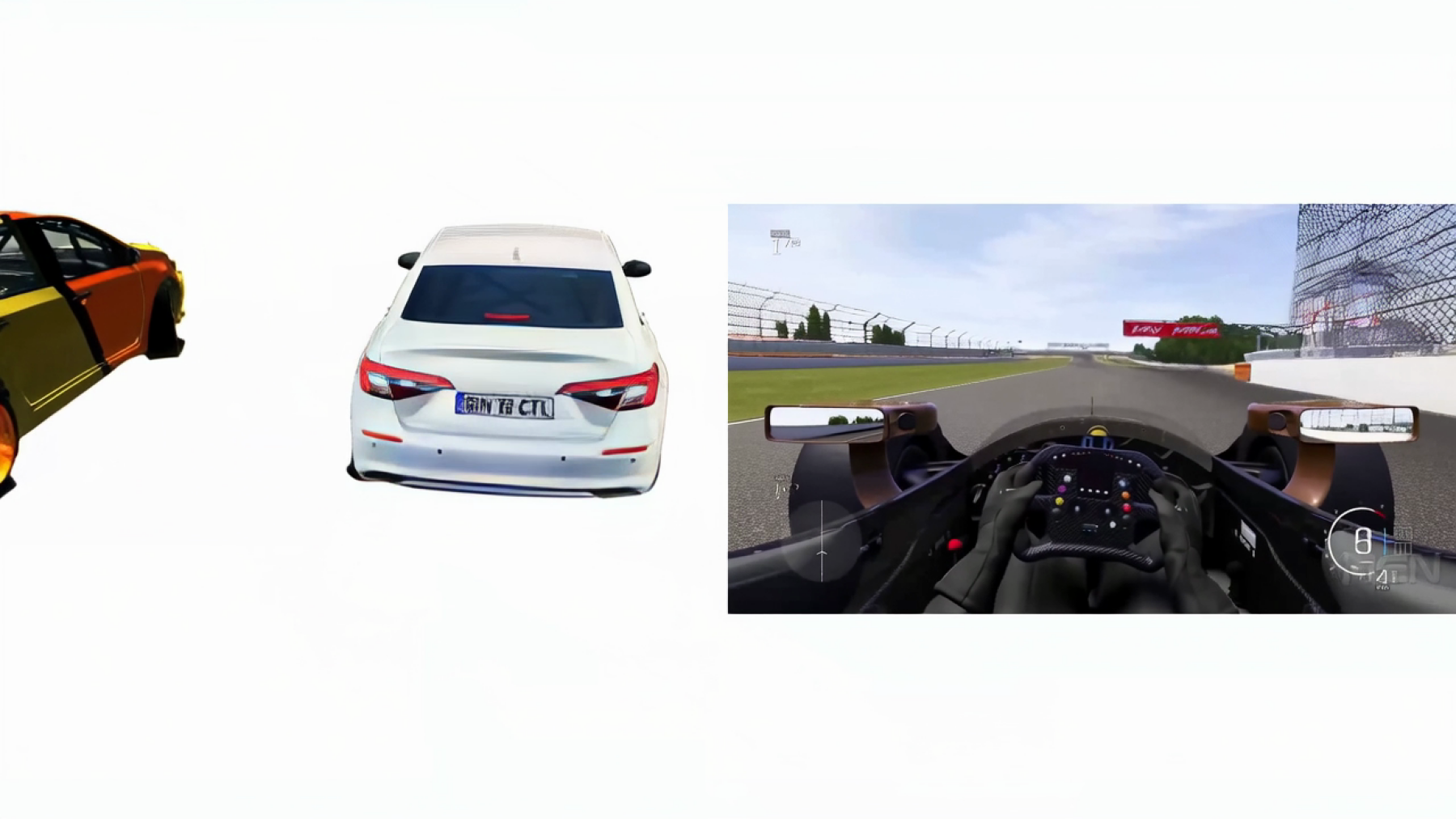}
                    };
                }%
                \hspace{0.3em}%
                \tikz[baseline={(current bounding box.center)}]{
                    \node[draw=blue, line width=2pt, inner sep=2pt] {
                        \includegraphics[width=0.18\linewidth]{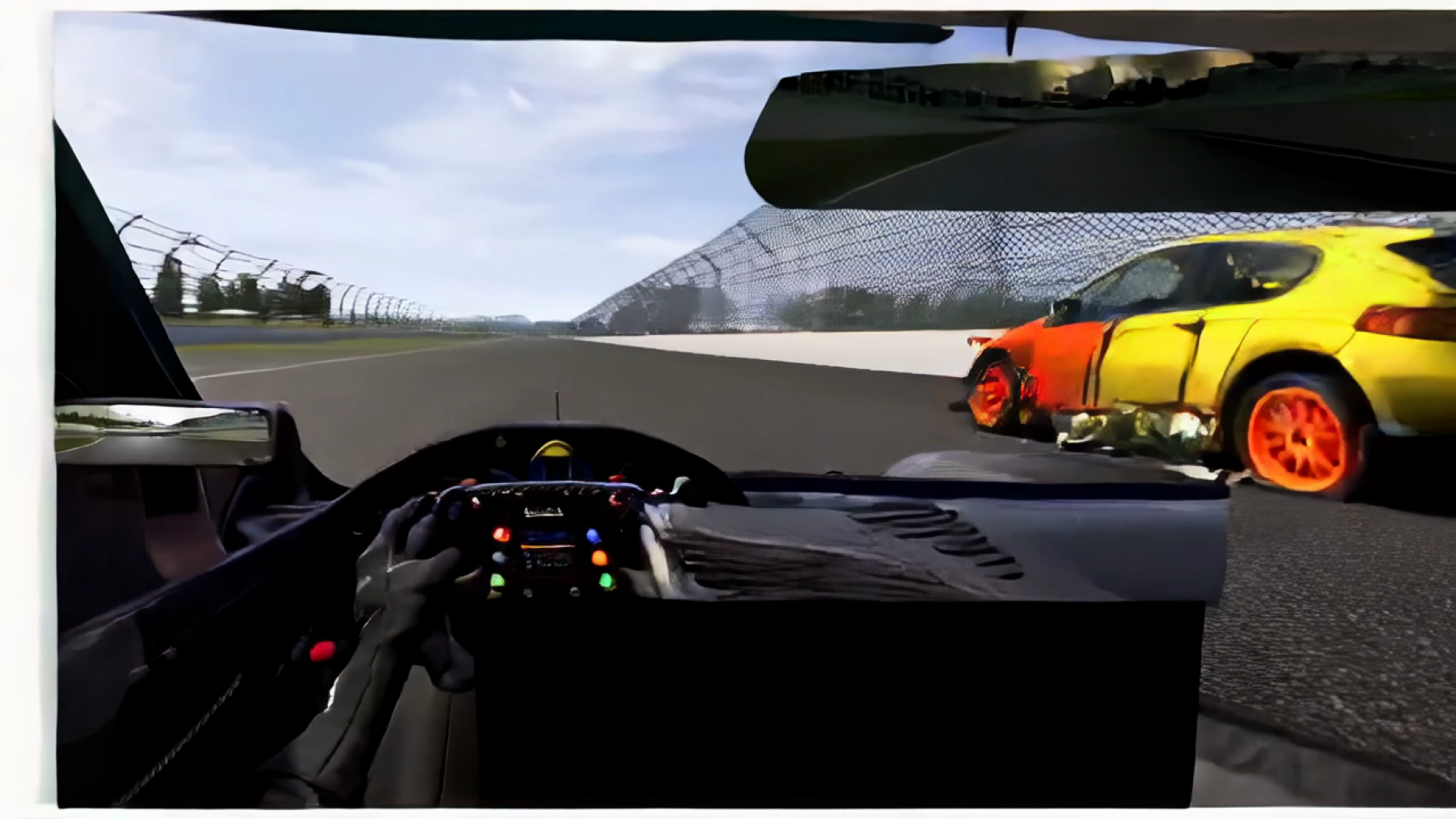}\hspace{0.3em}%
                        \includegraphics[width=0.18\linewidth]{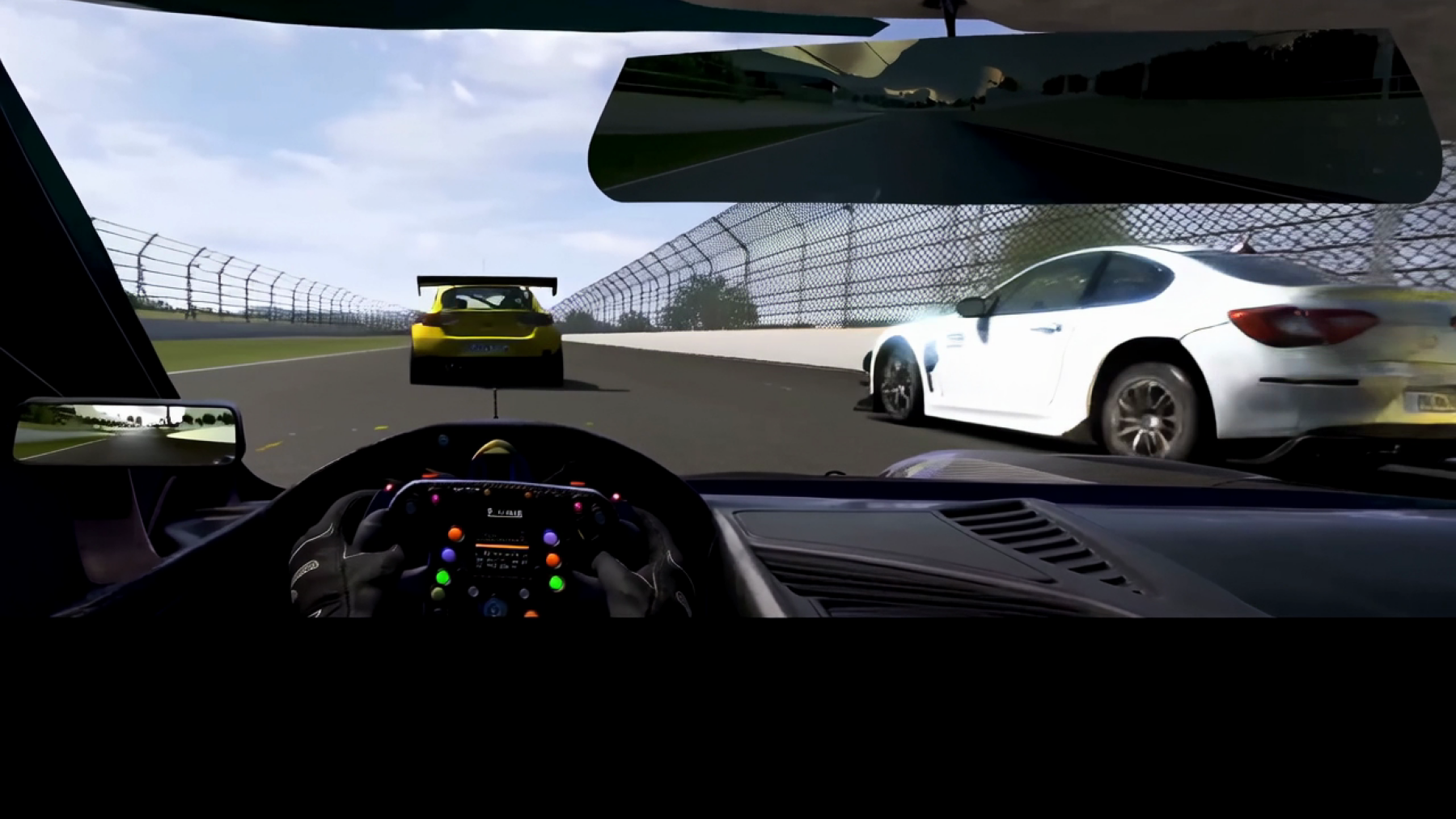}
                    };
                }%
            }
        };
    \end{tikzpicture}
    
    % Legend - side by side
    \vspace{5pt}
    \begin{tikzpicture}
        \node[inner sep=0pt] {
            \tikz{\draw[red, line width=2pt] (0,0) rectangle (0.4,0.25);}\hspace{0.5em}\small\textbf{Text Prompt}
            \hspace{2em}
            \tikz{\draw[blue, line width=2pt] (0,0) rectangle (0.4,0.25);}\hspace{0.5em}\small\textbf{$<$transition$>$ + Text Prompt}
        };
    \end{tikzpicture}

    \vspace{-6pt}
    
    \captionsetup{type=figure}

    \caption{In this figure, we illustrate a general yet under-explored observation: video generation models possess an innate ability to perform subject mixing via scene transitions from a mixed-subject first frame. As shown, the red-boxed results (without the transition phrase: $<$transition$>$) contrast with the blue-boxed results (with a carefully chosen $<$transition$>$ e.g., “The camera view suddenly zoom in to show") revealing significant differences in composition. However, this phenomenon faces three key limitations that hinder practical use: 1) The prompt engineering process for $<$transition$>$ is highly manual, time-consuming, and model/video-dependent. 2) Scene transitions are often unstable. 3) Object identity is often lost, resulting in changes in appearance or the disappearance of reference objects. }
    \label{fig:Transition_phrase}
    \vspace{-10pt}
\end{figure*}

To address this question, we investigate existing video generation models and uncover a previously overlooked yet fundamental capability: their innate ability to incorporate multiple reference concepts into the generation process without any architectural modifications or large-scale fine-tuning. As in Figure~\ref{fig:Transition_phrase}, \textit{standard video generation models have potentials to naturally embed visual concepts from multiple references into the first frame, functioning as a memory buffer, and then fuse them consistently through scene transitions during generation.} However, this ability is difficult to invoke directly: prompt engineering alone often leads to unstable outcomes and struggles to precisely preserve object identities or control visual composition. To transform this observation into a practical capability for multi-reference-based video customization, we propose a simple yet effective approach that reliably activates this innate ability. Using only 20-50 training video examples for lightweight training, our method enables the model to select visual concepts in the first frame and transition scenes coherently, achieving generalized multi-reference video customization. Importantly, this is done without modifying the model architecture or compromising the rich generative priors of the pre-trained video generation model.

To summarize our contributions in this work:
\begin{itemize}
\item We investigate the innate and general ability of video generation models, showing that the first frame not only serves as the spatiotemporal start of generation but also acts as a conceptual buffer, which enables multi-reference-based video content customization.
\item We develop a simple yet effective pipeline that leverages Vision-Language Models (VLMs) for high-quality training data curation and achieves invocation through just 20–50 examples via few-shot LoRA adaptation.
\item We evaluate and compare our proposed invocation add-on, \ours{}, with state-of-the-art video generation models across diverse applications, including filmmaking, generalized multi-object interaction, driving/aerial/underwater simulation, and robotic manipulation, etc.
\end{itemize}

\section{Related Work}
\paragraph{Video Generation and Video Content Customization.}
Video generation models~ \cite{blattmann2023stable, polyak2024movie, yang2024cogvideox, liu2024sora, wan2025wanopenadvancedlargescale, wiedemer2025videomodelszeroshotlearners} are a powerful class of generative models that synthesize videos conditioned on user-provided text prompts, primarily for creative content generation. Most state-of-the-art models are based on diffusion frameworks~\cite{rombach2022high, peebles2023scalable, chen2025inpainting} with U-Net-based denoising backbones, and more recently, Diffusion Transformers (DiT). A key limitation of these pre-trained models is their reliance solely on text prompts, which restricts controllability, particularly in real-world applications such as product demonstrations or simulations, where visual references are often required for precise customization.
To address this, recent works in video content customization~\cite{jiang2025vaceallinonevideocreation, fei2025skyreelsa2composevideodiffusion, liao2025character, chen2025multi, huang2025videomage, jiang2024videobooth, wei2024dreamvideo} explore extending video generation models to accept additional visual inputs. These approaches typically require: 1) architectural modifications, and 2) large-scale training on specialized datasets of customized videos. However, both requirements come with significant drawbacks. Architectural changes often compromise model efficiency and compatibility, while task-specific fine-tuning on limited domains may degrade the general knowledge acquired during large-scale pretraining, where data is more diverse, higher in quality, and broader in scope.

\vspace{-10pt}

\paragraph{The Innate Abilities of Pre-trained Generative Models}
Pre-trained generative models are typically trained on massive and diverse datasets, which endows them with general-purpose capabilities that often extend beyond their original design goals. These emergent properties, what we refer to as innate abilities, remain under-explored in current research, despite growing evidence of their utility in real-world applications. For example, recent work~\cite{huang2024context} demonstrates that with just a few fine-tuning samples and using Low-Rank Adaptation (LoRA)~\cite{hu2022lora} based fine-tuning, a pre-trained image generation model can be prompted to generate grid-aligned images with coherent content. In the video domain, study~\cite{wiedemer2025videomodelszeroshotlearners} havs shown that pre-trained image-to-video (I2V) models can perform various frame-level perception tasks such as edge detection, segmentation, and super-resolution, despite not being explicitly trained for them. Motivated by these findings, we explore whether similar innate abilities exist in pre-trained video generation models related to video content customization, and how they might be invoked through minimal adaptation, without architectural changes or large-scale retraining.

\vspace{-10pt}

\paragraph{Vision-language models (VLMs) for multimodal data curation.}
Vision-language models (VLMs) have become essential across a wide range of multi-modal applications, demonstrating strong capabilities in video understanding~\cite{feng2025video,li2025videohallu,liu2024mmbenchmultimodalmodelallaround}, video captioning~\cite{zhao2024distillingvisionlanguagemodelsmillions,Li_2025_CVPR,yang2023vid2seqlargescalepretrainingvisual}, and visual recognition tasks such as object detection and 3D scene understanding~\cite{madan2024revisiting,openai2024gpt4ocard, li2025selfrewardingvisionlanguagemodelreasoning,chen2024spatialvlmendowingvisionlanguagemodels, shi2024eagle}.
Recent progress in \textit{unified VLMs}, e.g., Gemini-2.5-Pro~\cite{Gemini2.5}, Qwen2.5-Omni~\cite{xu2025qwen25omnitechnicalreport}, GPT-4o~\cite{openai2024gpt4ocard} have further extended these capabilities by enabling seamless processing and generation across images, videos, and text within a shared representation space.
These models operate over a unified multimodal input space, capable of processing images, text instructions, and videos, and generating outputs across the same modalities, including images, textual responses, and videos~\cite{wang2024emu3nexttokenpredictionneed, yang2025baichuan2openlargescale,zhou2024transfusionpredicttokendiffuse, liu2023mitigating}.
In particular, unified VLMs demonstrate strong instruction-following capabilities for image editing tasks~\cite{wu2025qwenimagetechnicalreport} and video understanding tasks~\cite{feng2025video}, enabling us to leverage them for curating high-quality training data.

\section{Proposed Approach}
\label{sec:method}

\subsection{Pipeline Overview}

Overview of our pipeline is shown in Figure~\ref{fig:pipeline}. It consists of three main components: 1) Dataset Curation, which utilizes VLMs to generate high-quality paired training data inputs for corresponding videos. 2) Few-shot LoRA Adaptation, which invokes the model’s innate fusion and transition abilities, and 3) {Clean Customized Video Inference, which enables generalized multi-reference video generation.

\begin{figure*}[!t]
    \centering
    \includegraphics[width=.999\linewidth]{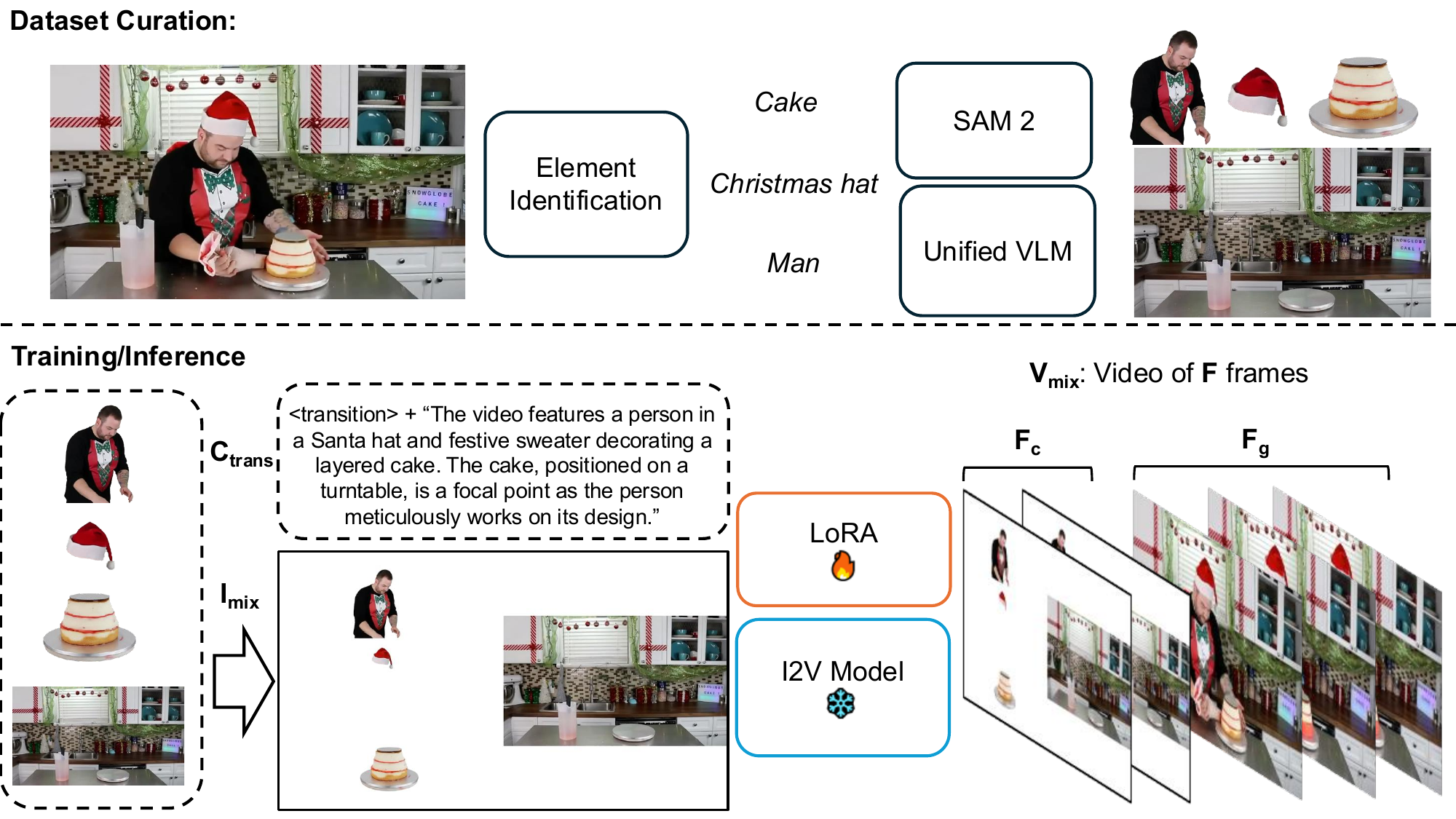}

    \vspace{-8pt}

    \caption{The overview of our proposed pipeline \ours{}, consists of 1) Dataset Curation for getting the high quality finetuning data from existing videos, 2)  Few-shot LoRA Adaptation for training/inference to invoke the I2V model's innate ability in fusing the subjects in the first frame and perform a scene trasition to generate a video $V_{mix}$ following subjects in the first mixing frame $I_{mix}$  and the text prompt. }
    \label{fig:pipeline}
    \vspace{-10pt}
\end{figure*}

\subsection{Dataset Curation}
Previous subject mixing models such as SkyReels-A2~\cite{fei2025skyreelsa2composevideodiffusion}, VACE~\cite{jiang2025vaceallinonevideocreation} modified model architectures to take in multiple reference images and uses millions of training samples to train models to mix subjects, where the subjects are mainly with humans and does not generalize to other applications such as autonomous driving, robot manipulation.
Instead of modifying the models' architecture and using millions of training data to learn subject mixing, we leverage the models' pre-trained abilities to learn subject mixing without modifying its architecture.
To do so, we need to prepare a dataset where the input aligns with the base model (Wan2.2~\cite{wan2025wanopenadvancedlargescale})'s input, which is an image and a text prompt.

\vspace{-10pt}

\paragraph{Training Data Selection.} We gather two thousand videos from varies sources include Veo 3 showcases~\cite{wiedemer2025videomodelszeroshotlearners}, HOIGen-1M~\cite{liu2025hoigen1mlargescaledatasethumanobject} and licensed short videos. 
Each of the videos is ranging from six to twenty seconds long, we crop all the videos to keep the first 81 frames.
Next, we carefully go over a subset of the videos, and pick the high quality ones that have clear interaction or combinations of different elements, with clear and complete boundaries that can be easily segmented.
Then for each video we think it is high quality, we write down the elements we want to segment out from the video in text form. For example, in a video where a man wearing a Chrismas hat is making a cake, we write down the element names \textit{the man, cake, Chrismas hat}, which will be used to segment out later.
We use the first frame of the selected videos to extract elements and process to be the input for the training data.
We sample a total of 50 carefully selected videos with human annotations of objects to be segmented out.

\vspace{-10pt}

\paragraph{Element Extraction using unified VLMs.} Given an image $I$, and a set of elements in textual representation $O$, our goal is to use a unified VLM to (1) recognize and extract and generate each of the individual $O$ from $I$, (2) remove all $O$ from $I$ and generate a complete background of $I$.
Specifically, we prompt Gemini-2.5-Pro~\cite{Gemini2.5} to do this process. 
At them end, we use SAM 2~\cite{ravi2024sam2segmentimages} to remove the white backgrounds of the individually extracted or generated elements and keep an RGBA layer for each $O$, where only the element has RGB layer.
Given the set of elements $O$ in RGBA and a background, we combine them into a single image, with the background always on the right and the elements always on the left with automatic resize to fit the elements.

\vspace{-10pt}

\paragraph{Prompt generation using VLMs.} Given a set of RGBA elements, and a ground truth video, we prompt Gemini-2.5-Pro to generate a prompt that not only describe the video, but also focuses on the appearance and interaction of these elements. Thus, we get a final training dataset with combined images with their respective background and elements and a caption that describes how these elements fit into the video and background.

\subsection{Few-shot LoRA Adaptation}
As shown in Figure~\ref{fig:pipeline}, with only 20–50 training examples, we apply a simple LoRA-based few-shot adaptation to a pre-trained model, enabling it to learn subject mixing by fusing subjects in the first frame and performing a scene transition. Given a pre-trained I2V model $g_{\theta}$, input image $I$, and text prompt $C$, the standard generation process is: $V = g_{\theta}(I, C)$, where $V$ is a video of $F$ frames. In our adapted pipeline: The input image is replaced by a subject-mixed image $I_{mix}$. The prompt is modified by appending a special identifier \cite{ruiz2023dreambooth} to indicate a transition $C_{trans} = <transition> + C$. A LoRA-based weight update $\Delta \theta$ is applied to the model.  LoRA learns an updated weight matrix $\theta$ defined as $\theta$ = $\theta$ + $\Delta \theta$, where $\theta$ represents the original pre-trained weights and $\Delta \theta$ is a learnable low-rank update. Instead of learning a full-rank matrix of size $d\times k$, LoRA factorizes the update as: $\theta =\theta + \alpha A B$, where $\alpha$ is a scaling factor, $A \in R^{d\times r}$ , $B \in R^{r\times k}$, and $r$ is a low-rank dimension much smaller than both $d$ and $k$.
The target output video $V_{mix}$ exhibits both subject mixing and a temporal scene transition. Specifically, the video has frame length $F = \{F_{c} , F_{g}\}$ , where: $F_{c}$ represents the temporal compression frames, namely the temporal compression ratio (e.g., $F_{c} =4$ in the Wan2.2), $F_{g}$ contains the generated subject-mixed content, following a sudden transition after frames $F_{c}$. Thus, the adapted video generation process becomes: $V_{mix} = g_{\theta + \Delta \theta}(I_{mix}, C_{trans})$. 

\subsection{Clean Customized Video Inference}
During the inference, to generate a content customized video, the processing is very simple, since the model will generate the  $V_{mix}$ has frame length $F = \{F_{c} , F_{g}\}$ with both subject mixing and a temporal scene transition. The users can easily cut off the first $F_{c}$ frames to get the clean subject mixing videos. As an example, in Wan2.2 as the base model, to generate a $F =81$ frame video, the first $F_{c} =4$ frames will be discarded and the rest $F_{g}= 77$ frames will be the clean customized content videos.
% \section{Evaluation and Result}

\section{Experimental Results}
\label{sec:result}
The results and comparisons are best viewed through video examples. We include video results in our project webpage.

\subsection{Datasets and Implementation Details}

\paragraph{Few-shot Training Data.}  We carefully selected 50 videos featuring object-object, human-human, or human-object interactions with clearly defined boundaries suitable for segmentation. These were curated from a pool of 2,000 videos sampled from HOIGen-1M~\cite{liu2025hoigen1mlargescaledatasethumanobject}, licensed video clips, and Veo 3 samples~\cite{wiedemer2025videomodelszeroshotlearners}.
Each curated video contains 81 frames. The dataset spans four main categories: human-object interaction (60\%), human-human interaction (14\%), element insertion (20\%), and robot manipulation (6\%). Details of the curated training set are provided in the Supplementary Materials.

\vspace{-10pt}

\paragraph{LoRA Adaptation Training Details} We use Wan2.2-I2V-A14B~\cite{wan2025wanopenadvancedlargescale} as our base model. For LoRA adaptation, we train with a LoRA rank of 128 and introduce a unique transition phrase (An arbitrary phrase, as long as it is unique, following a similar intuition to DreamBooth \cite{ruiz2023dreambooth}), $<$transition$>$: “ad23r2 the camera view suddenly changes.” This phrase serves as a prompt to trigger the model’s innate ability for subject selection and scene transition from the first frame. Since Wan2.2-I2V-A14B employs two separate denoising transformers for low- and high-noise regimes, we train each independently for 5 hours using 2 NVIDIA H200 GPUs, with a batch size of 4.

\begin{table*}[h]
\centering
\begin{tabular}{lccccc}

\hline
\hline
\textbf{Model} & \textbf{Overall Quality $\uparrow$} & \textbf{Object Identity $\uparrow$} & \textbf{Scene Identity $\uparrow$} & \textbf{Avg. Rank $\downarrow$} & \textbf{\% Ranked 1st $\uparrow$} \\
\hline
Wan2.2-I2V-A14B  & 2.09 & 3.32 & 3.01 & 3.27 & 3.4\% \\
SkyReels-A2 & 2.34 & 2.89 & 3.43 & 3.02 & 4.3\% \\
VACE & 3.00 & 3.50 & 3.66 & 2.50 & 11.1\% \\
Ours  (\ours{}) & \textbf{4.28} & \textbf{4.53} & \textbf{4.58} & \textbf{1.21} & \textbf{81.2\%} \\
\hline
\hline

\end{tabular}

\vspace{-7pt}

\captionsetup{font=small}
\caption{User Study Results: We report ratings and rankings from 200 annotations across 40 users. \ours{} consistently outperforms all baseline models across evaluation aspects, despite being trained on only 50 examples, demonstrating significantly better generalization across diverse application scenarios.}

\label{tab:model_eval}

\vspace{-13pt}

\end{table*}

\subsection{Evaluation Strategy}
In this section, we outline our evaluation strategy to demonstrate the effectiveness of our proposed model and its performance relative to baseline methods. 
% We describe the curation process of our test set, the range of application domains included in our evaluation, and the baseline models selected for comparison.

\vspace{-11pt}

\paragraph{Test set spanning diverse application scenarios.} To rigorously evaluate the effectiveness and generalization of our approach for video content customization, we curate a test set of 50 reference sets, each composed of materials from different sources, spanning diverse applications such as robot manipulation, filmmaking, aerial/driving/underwater simulation, and product demonstrations. Compared to prior works~\cite{jiang2025vaceallinonevideocreation, fei2025skyreelsa2composevideodiffusion}, our test set offers two key advantages: (1) it covers a broader range of real-world customization scenarios beyond the typical human-human or human-object interactions; and (2) it supports up to 5 reference subjects, exceeding the 3-subject limit (e.g., object1, object2, scene) seen in previous works.

\vspace{-11pt}

\paragraph{Baseline Models}
We compare our method against three strong baselines built on the Wan architecture with 14 billion parameters: Wan2.2-14B-I2V~\cite{wan2025wanopenadvancedlargescale}, VACE~\cite{jiang2025vaceallinonevideocreation}, and SkyReels-A2~\cite{fei2025skyreelsa2composevideodiffusion}. Wan2.2-14B-I2V is our base I2V model, to which we apply our lightweight adaptation for invoking its innate subject mixing and scene transition capabilities. As shown in Figures \ref{fig:bear}, \ref{fig:wukong} and \ref{fig:rocket}, our adaptation significantly enhances its performance in video content customization while preserving the original generation quality (Figure \ref{fig:base_comp}). VACE and SkyReels-A2 represent state-of-the-art video customization models trained on millions of high-quality examples. Despite their scale, we demonstrate that our method can outperform both using only 50 training videos across a wide range of scenarios.

\subsection{Qualitative Comparison}
In this section, we qualitatively compare our proposed method with baseline models across diverse scenarios.

\vspace{-10pt}

\paragraph{Comparison with the Base Model.}
We first compare our adapted model with the base model (Wan2.2-I2V-A14B) to demonstrate the effectiveness of our approach in video content customization, as shown in Figures~\ref{fig:bear}, \ref{fig:wukong}, and \ref{fig:rocket}. For the base model, we use the mixed image input along with a fixed transition phrase ($<$transition$>$) that empirically performs best. As observed, the base model often animates elements independently, and the referred objects tend to disappear post-transition. In contrast, our adapted model consistently preserves object identities and performs coherent scene transitions, indicating a substantial improvement in handling reference-based video customization.

\vspace{-10pt}

\paragraph{Preserving Pre-trained Knowledge in the Base Model.}
As demonstrated earlier, our add-on significantly enhances the base model’s performance in reference-based video customization. Crucially, it also preserves the base model’s original pre-trained knowledge. Since our method is designed to invoke the model’s innate capabilities rather than overwrite them, it retains the core generative priors embedded through pre-training. This preservation is illustrated in Figure~\ref{fig:base_comp}. In the rare successful cases where the base model maintains all object identities and executes a coherent scene transition, our results closely mirror the base output, in this instance, replicating the motion and positioning of the wingsuit performer and the car. This fidelity underscores an important advantage: our method enables customization without compromising the valuable general knowledge encoded during pre-training. Given that post-training data for customization is often narrower in scope and lower in quality than pre-training corpora, it is critical for video content customization approaches to integrate new reference inputs while preserving the strengths of the base model. Our add-on achieves exactly this, offering a lightweight yet effective pathway to transform general-purpose I2V models into powerful, user-controllable video customization systems.

\begin{figure}[!t]
    \centering
    \includegraphics[width=1\linewidth]{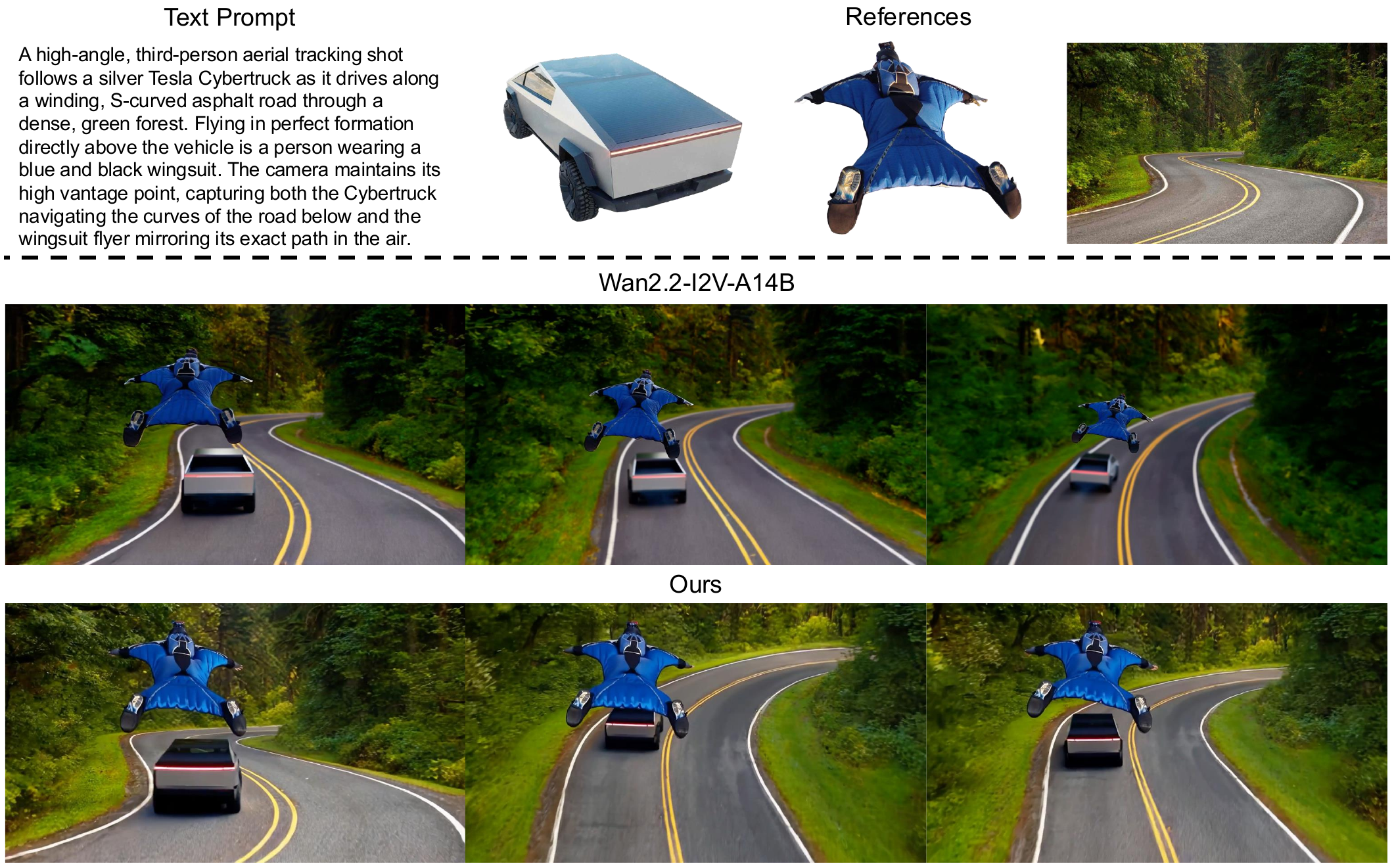}

    \vspace{-10pt}

    \caption{As shown in the figure, in rare cases where the base model Wan2.2-I2V-A14B successfully performs a scene transition while preserving all reference object identities, the output closely resembles ours. This demonstrates that our add-on approach effectively retains the base model’s pre-trained generative capabilities. }

    \label{fig:base_comp}
    \vspace{-15pt}
\end{figure}

\vspace{-10pt}

\paragraph{Comparison with State-of-the-Art Video Customization Baselines.} We compare our method with two Wan-based state-of-the-art baselines: VACE and SkyReels-A2. 1) As shown in Figures~\ref{fig:bear} and~\ref{fig:rocket}, both baselines are trained on millions of videos for specific customization tasks involving the composition of three elements: human, object, and scene. However, this design leads to overfitting, limiting their generalization to novel application scenarios. In contrast, our method, trained on just 50 examples, activates the pre-trained base model's innate capabilities while preserving its general knowledge, enabling superior performance across diverse use cases. 2) Figure~\ref{fig:wukong} highlights a key limitation of these baselines: their architecture only supports up to three reference inputs. As a result, they fail to handle scenarios with five references. In contrast, our approach has no such architectural constraint, as our multi-reference capability stems from the the utilization of the first frame as conceptual memory buffer, not architectural modification.

\begin{figure*}[t]
    \centering
    \includegraphics[width=1\linewidth]{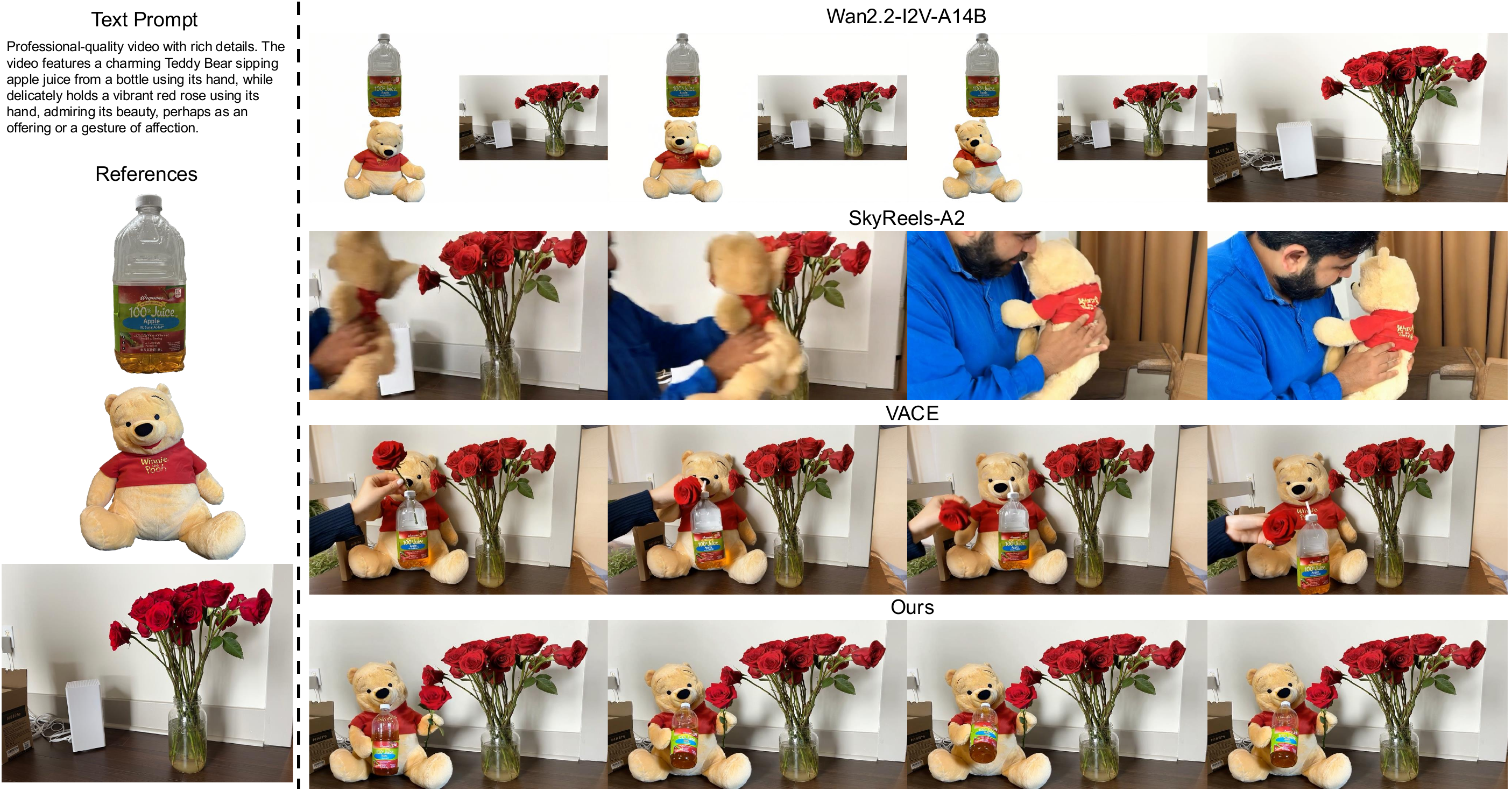}  

    \vspace{-11pt}

    \caption{Qualitative comparison with baseline methods. This test scenario involves generalized multi-object interactions. As shown in the figure, our method best preserves the identities of input objects and the scene, while generating a customized video with coherent motion that aligns with the text prompt description.}

    \label{fig:bear}
    \vspace{-7pt}
\end{figure*}

\begin{figure*}[t]
    \centering
    \includegraphics[width=1\linewidth]{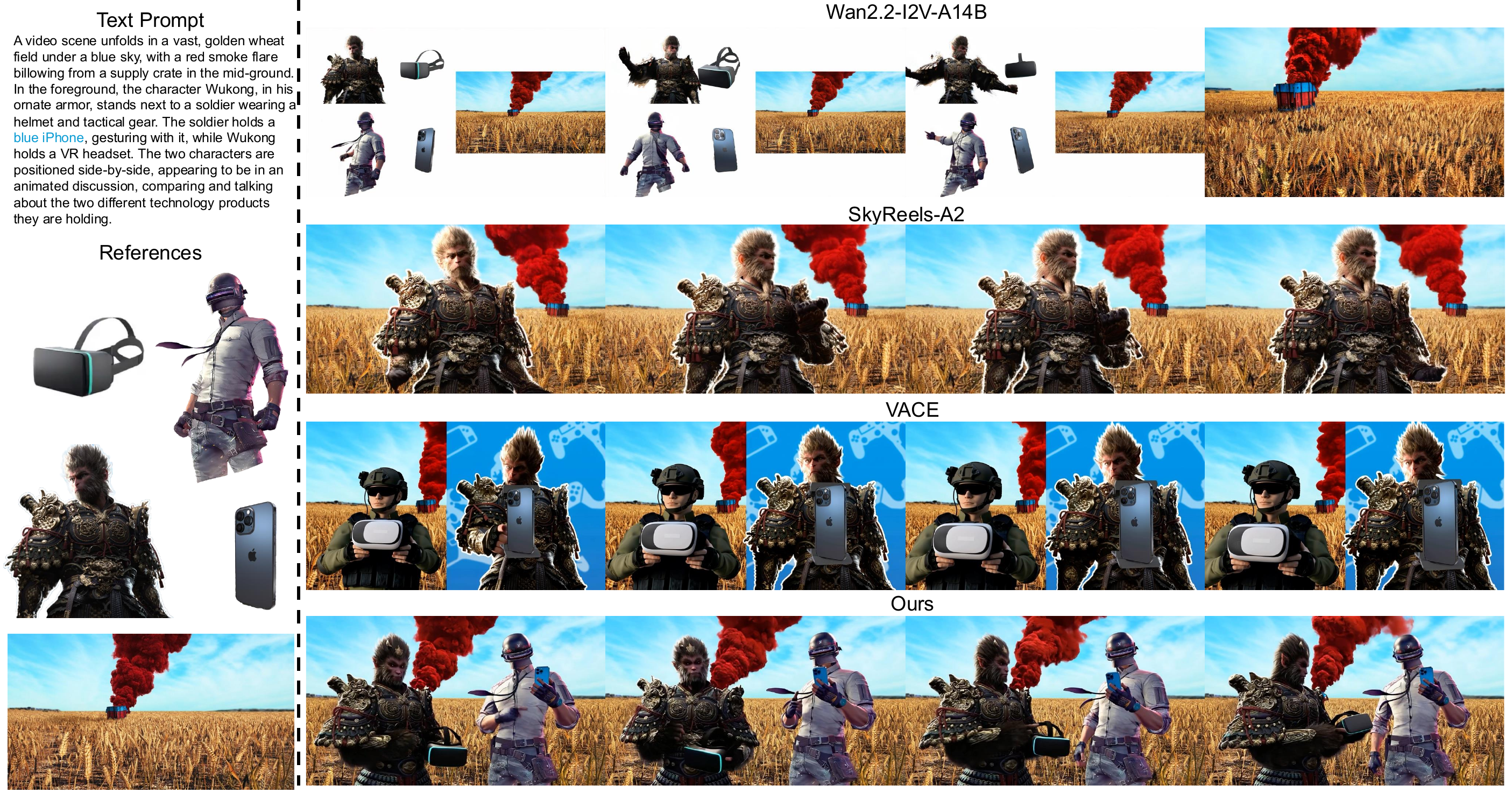}
    
    \vspace{-10pt}

    \caption{Qualitative comparison with baseline methods. This scenario evaluates performance with an excessive number of references, five in total (four objects and one scene). VACE and SkyReels-A2, due to their architecture-based limitations, support only up to three references and fail to include all four reference objects in the generated video. In contrast, our model successfully fuses all four objects into a coherent, customized video with natural interactions. Notably, our model also enables precise selection via text prompt (e.g., \textcolor{blue}{blue iPhone}), preserving key visual traits such as the triple-camera design while modifying appearance (e.g., changing the color to blue).}

    \label{fig:wukong}
    \vspace{-7pt}
\end{figure*}

\begin{figure*}[t]
    \centering
    \includegraphics[width=1\linewidth]{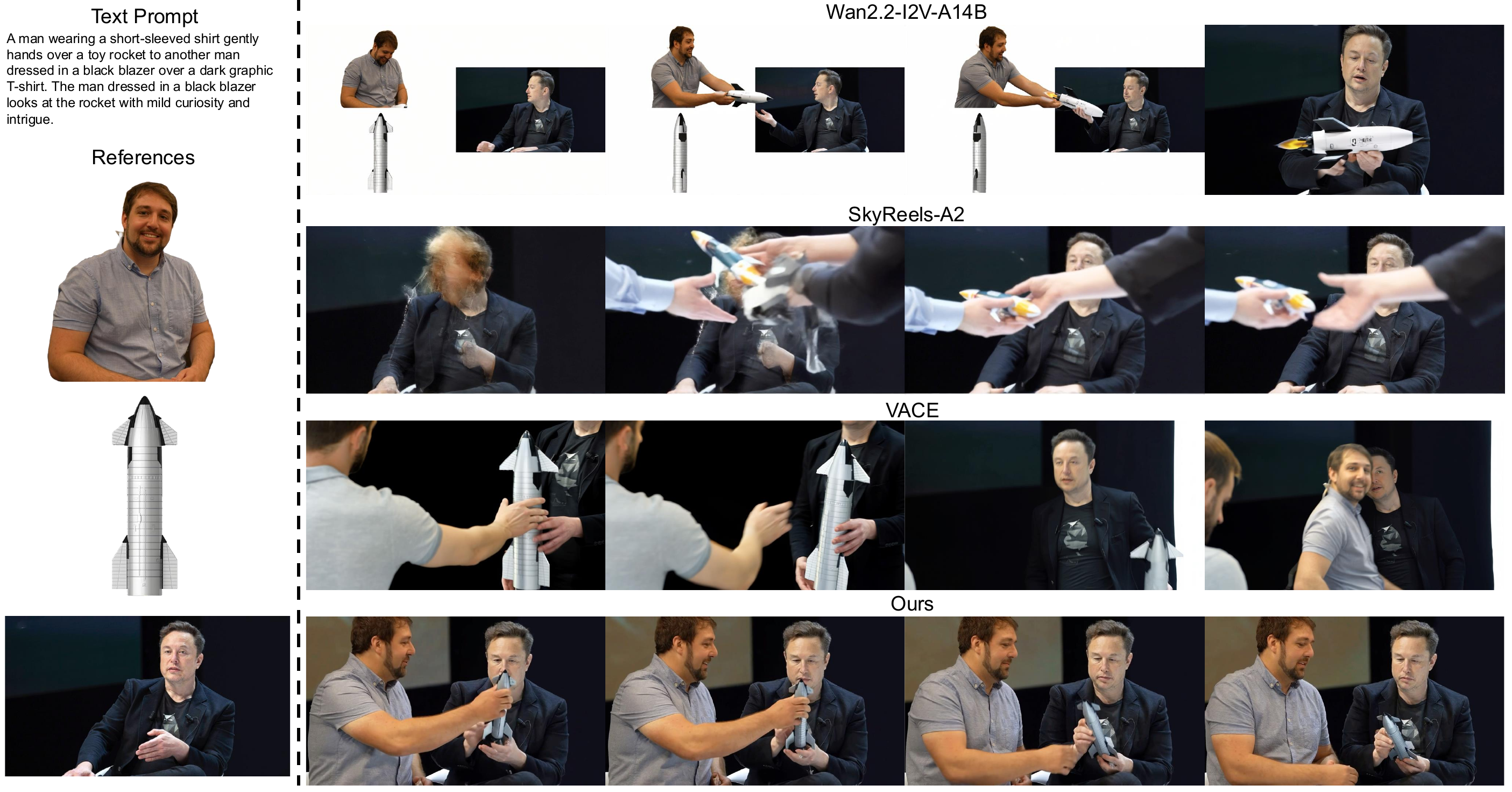}  
    \vspace{-20pt}

    \caption{Qualitative comparison with baseline methods.
This scenario evaluates generalized human-object interactions involving multiple humans. While both VACE and SkyReels-A2 excel in customized single human-object video generation, they struggle in more complex multi-human scenarios where interactions are mediated by shared objects. In such cases, both baselines fail to maintain object integrity and coherent interaction. In contrast, our method reliably generates consistent videos with preserved object identity and realistic multi-human interactions.}

    \label{fig:rocket}
    \vspace{-7pt}
\end{figure*}

\subsection{Quantitative Comparison}
To quantitatively evaluate our method and compare it with baselines, we conducted a user study on the full test set across all methods. Details of the user study setup are provided in the Supplementary Materials. We built an online interface where users are shown the prompt and a reference image containing all foreground objects and background context. The system then displays four generated videos in randomized order.
Users were asked to: 1) \textbf{Rank} the four videos from best (1st place) to worst (4th place).
2) Rate each video on three aspects: \textbf{Object Identity: } with a score range from 1 (worst) to 5 (best), \textit{How well are foreground object identities preserved from the reference image?}. \textbf{Scene Identity: } with a score range from 1 (worst) to 5 (best), \textit{How accurately is the background retained?} and \textbf{Overall Quality: }  with a score range from 1 (worst) to 5 (best), \textit{How realistic and coherent is the video overall?}.
Table~\ref{tab:model_eval} presents the results. Our method (FFGo) outperformed all baselines across every metric, rank, object identity, scene identity, and overall quality, despite using only a lightweight adaptation. Notably, over 80\% of users selected our results as their top choice, indicating strong alignment with real user preferences.
Importantly, \ours{} transforms the base model Wan2.2-I2V-A14B from the lowest-performing baseline to the top performer in user evaluations. This highlights the strength of our add-on approach, which achieves state-of-the-art customization performance without architectural changes or large-scale training.

\section{Limitations}
While our adaptation effectively invokes the innate ability of pre-trained I2V models for video content customization via the first frame, several limitations remain. Although it is theoretically possible to incorporate an arbitrary number of reference subjects in the first frame, in practice, increasing the number of subjects reduces the resolution available to each, making identity preservation more difficult.
Another challenge is selective control: as the number of reference subjects grows, it becomes harder to reliably target specific objects using only text prompts. Empirically, we find our method performs well up to four subjects plus a reference scene (five references in total), beyond which identity preservation and prompt-based selection degrade.
We believe these limitations are not fundamental and can be addressed through engineering improvements. For example, using multiple start frames as an extended conceptual memory buffer could allow for higher-capacity reference encoding. We leave such enhancements to future work.

\section{Conclusions}
In this work, we propose a fundamentally different perspective on the role of the first frame in video generation models. Contrary to the standard view that treats it merely as the spatiotemporal starting point of an animation, we show that the first frame functions as a conceptual memory buffer, capable of storing and fusing disjoint reference subjects for downstream generation. Building on this insight, we introduce a lightweight, add-on method to invoke this overlooked innate ability for video content customization. Without modifying the model architecture or requiring large-scale finetuning, our few-shot adaptation turns a base video generation model into a state-of-the-art video customization system. We demonstrate strong performance across a wide range of real-world scenarios and validate our approach through a comprehensive user study, showing clear alignment with user preferences.

\clearpage

\paragraph{Acknowledgment:} We greatly acknowledge NSF’s support under awards OISE 2020624 and BCS 2318255.

{
    \small
    \bibliographystyle{ieeenat_fullname}
    \bibliography{main}
}

% % WARNING: do not forget to delete the supplementary pages from your submission 
\clearpage

\maketitlesupplementary
\thispagestyle{empty}
\appendix

\tableofcontents

\section{Video Results}
Please refer to our project page: \url{http://firstframego.github.io} for video results, which clearly demonstrate the effectiveness of our method and its comparison with baseline models.

\section{Comparison with Two-Stage Baselines}
Fig.~\ref{fig:combined} shows a comparison with a representative two-stage baseline: First composing an image layout by an image composition model like MS-Diffusion \cite{wang2024ms} and then using an I2V model for animation. Our approach offers two key advantages: 1) Our method is fully end-to-end; 2) Temporal Control via Text Prompts: {More importantly, our method preserves fine-grained temporal control. For example, prompts like “a shark joins the party later” are faithfully realized. In contrast, two-stage pipelines lose this capability, as the fixed first frame dictates the entire spatial layout, limiting temporal flexibility.

\begin{figure}[t!]
    \centering
    
    \includegraphics[width=1.0\linewidth]{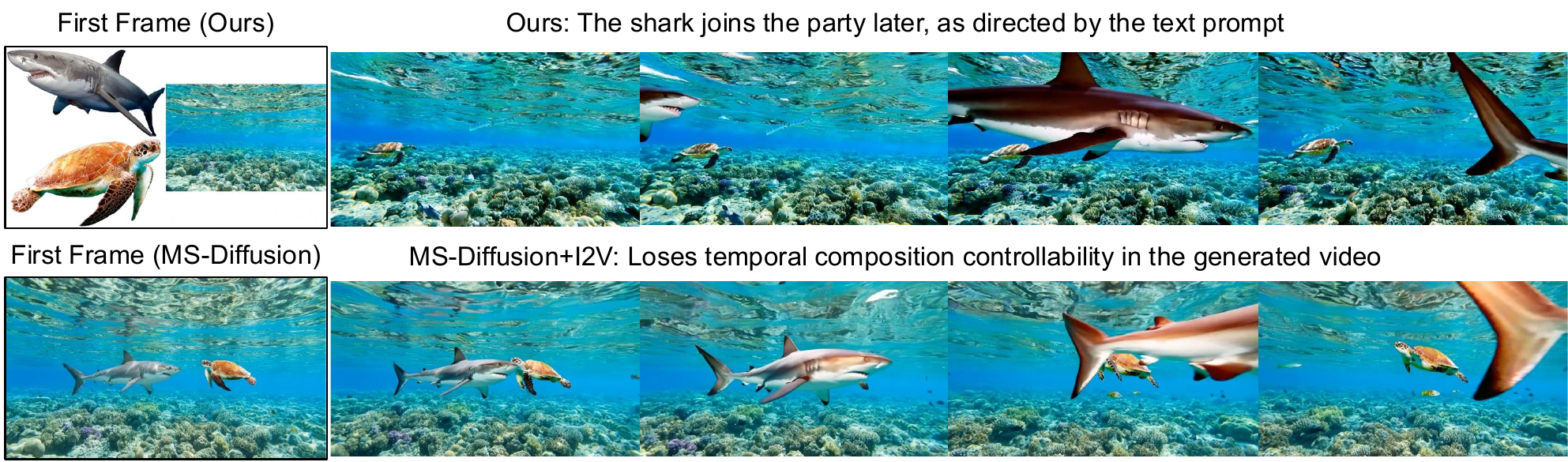} 
    % \vspace{-16pt}

    \small
    \vspace{-10pt}
    \caption{Comparison with MS-Diffusion + I2V.}

    \label{fig:combined}
    \vspace{-13pt}
\end{figure}

\section{Details about Training and Testing Set}
\subsection{Training Dataset Curation Details}

Our training corpus is sourced from three datasets: one randomly selected folder from HOIGen-1M \cite{liu2025hoigen1mlargescaledatasethumanobject} ($\approx$ 2,000 clips), all five Veo 3 demonstration videos \cite{wiedemer2025videomodelszeroshotlearners}, and 200 licensed short videos. This yields 2,205 candidate clips.

We manually curate the data to select videos with clearly separable foregrounds, humans or manipulable objects, set against uncluttered backgrounds. Only clips depicting cleanly segmentable single- or multi-object interactions are retained.

This filtering results in 50 high-quality training examples (Figure~\ref{fig:train_data}), distributed across four scene types: human–object interaction (60\%), human–human interaction (14\%), element insertion (20\%), and robot manipulation (6\%).

\paragraph{Training Data Processing.}
After curation, all clips are standardized to 81 frames for consistent training. From each video, we extract the first frame as a reference image and manually tag all foreground entities of interest, e.g., \emph{cake}, \emph{party hat}, \emph{male presenter}, \emph{mouse}. Using a prompt-to-prompt workflow with Gemini-2.5-Pro, we then perform:

\begin{itemize}
\item \textbf{Object Extraction:} Apply Prompt~\ref{fig:object_extraction_prompt} to generate high-fidelity renditions of each tagged entity, preserving their original appearance and scale. We refine results using SAM 2 or Adobe Photoshop to isolate each object as an RGBA layer.
\item \textbf{Background Cleanup:} Use Prompt~\ref{fig:object_removal_prompt} to produce a clean companion image with all tagged objects removed, yielding a pristine background plate.
\end{itemize}

This paired set of object cut-outs and object-free backgrounds forms the compositional basis of the training first frame.

\paragraph{Caption Generation.}
We use Gemini-2.5-Pro to generate rich, element-aware captions for each training sample, based on the individual object cut-outs, clean background plate, and the full 81-frame video. These inputs are paired with a structured prompt template (Fig.~\ref{fig:training_caption_generation_prompt}) to ensure consistency and relevance.

\paragraph{Element Composition for First Frame.}
For each training clip, we synthesize a 1280×720 reference canvas: all foreground cut-outs are vertically tiled on the left half, while the clean background is centered on the right (see Fig.~\ref{fig:train_data}). This composite serves as both the conditioning input and the initial frame, guiding the video generation model to blend the elements into a coherent sequence.

\subsection{Test Set Curation}
We manually curated a diverse test set of foreground objects and backgrounds from our self-collected images. Each object was segmented using SAM 2 or Adobe Photoshop and saved as an RGBA cut-out. These cut-outs were then composited with their respective backgrounds on a 1280×720 canvas, following the same layout used in training.

For each object-background pair, we drafted an initial prompt and refined it using Gemini-2.5-Pro with the template shown in Fig.~\ref{fig:test_video_prompt_enhanced}. This process produced 50 high-quality prompts paired with composite reference images, forming our final test set.

\begin{figure}[ht]
  \centering
  \begin{tcolorbox}[width=\columnwidth,      % <-- fits exactly one column
                    colback=yellow!10!white,
                    colframe=yellow!50!black,
                    title={\large\bfseries Object Extraction Task Prompt Template},
                    fonttitle=\bfseries,
                    sharp corners,
                    boxrule=0.8pt,
                    left=3mm, right=3mm, top=2mm, bottom=2mm,
                    enhanced]

    \textbf{Prompt} – Given the input image, extract the subset
    \texttt{\{IDENTIFIED OBJECT\}} (i.e., only the specified foreground objects)—
    return them \emph{alone} with \textbf{no resizing, compression, or
    background} so the output resolution exactly matches the original image.
  \end{tcolorbox}
  \caption{Prompt for extracting identified foreground objects using a unified VLM.}
  \label{fig:object_extraction_prompt}
\end{figure}

\begin{figure}[ht]
  \centering
  \begin{tcolorbox}[width=\columnwidth,   % fits exactly one column
                    colback=yellow!10!white,
                    colframe=yellow!50!black,
                    title={\large\bfseries  Object Removal Task Prompt Template},
                    fonttitle=\bfseries,
                    sharp corners,
                    boxrule=0.8pt,
                    left=3mm, right=3mm, top=2mm, bottom=2mm,
                    enhanced]  % borders + neat line breaks

\textbf{Prompt} – Given the input image, \textbf{remove} the subset
\texttt{\{IDENTIFIED OBJECTS\}} entirely.  Return the edited image \emph{only}—it must
preserve the source resolution (no scaling or compression) and contain neither
the specified objects nor any artifacts of their removal.

  \end{tcolorbox}

  \caption{Prompt and specifications for the object removal task.}
  \label{fig:object_removal_prompt}
\end{figure}

\begin{figure}[t]
    \centering
    
    \includegraphics[width=.999\linewidth]{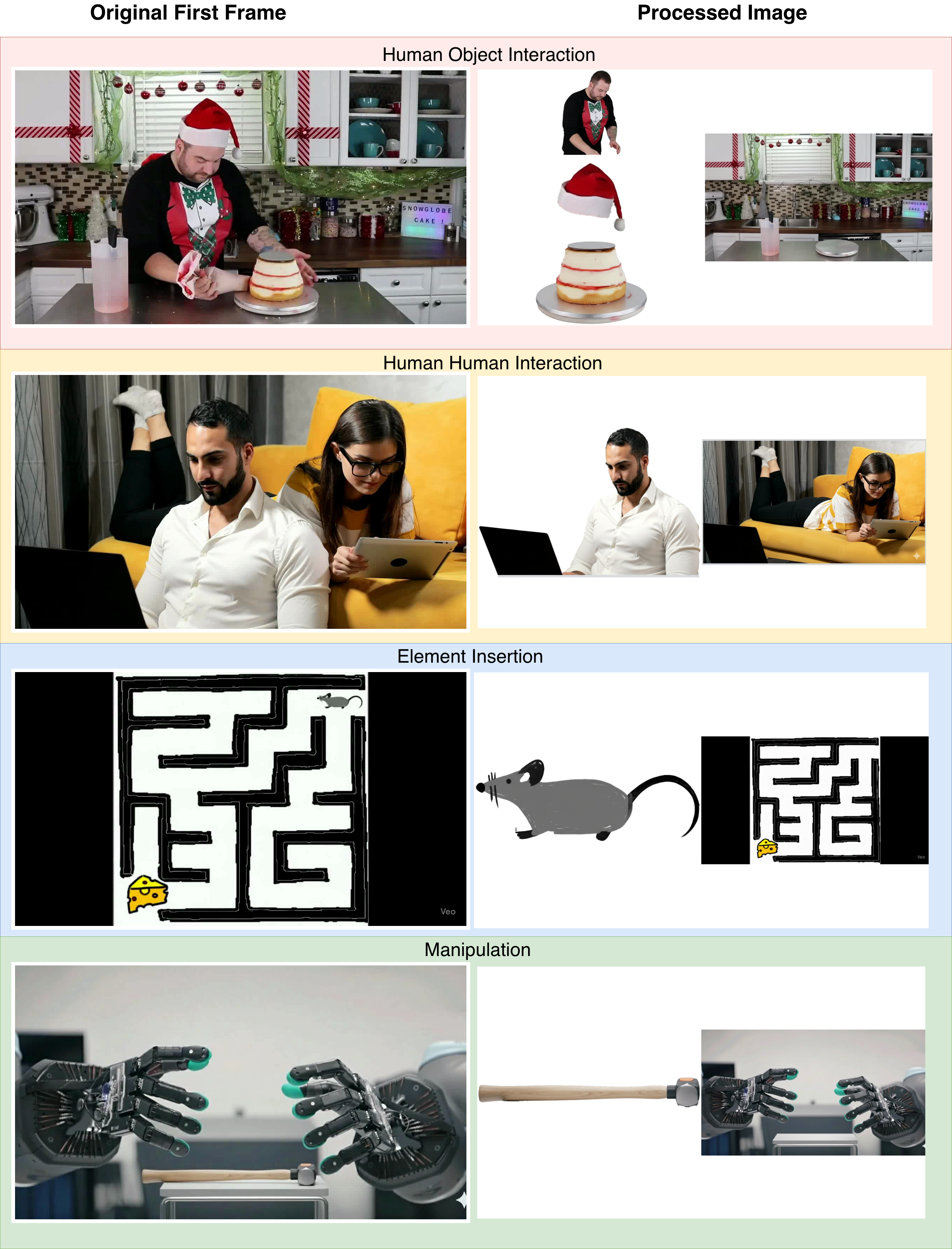} 
    \caption{Our training dataset comprises four categories: human–object interaction (60\%), human–human interaction (14\%), element insertion (20\%), and robot manipulation (6\%).}
    \label{fig:train_data}
    \vspace{-8pt}
\end{figure}

\begin{figure*}[t]
  \centering
  \begin{tcolorbox}[
      width=\textwidth,
      enhanced,
      % breakable,
      colback=yellow!6!white,
      colframe=yellow!60!black,
      colbacktitle=yellow!60!black,
      coltitle=white,
      title={\Large\bfseries Training Data Prompt Generation Prompt Template},
      fonttitle=\bfseries,
      sharp corners,
      boxrule=0.9pt,
      left=4mm, right=4mm, top=2mm, bottom=2mm
  ]
  \small  % compact text for neat fit

  %-----------------------------------------------------
  \textbf{Task Description}\\[4pt]
  You are given a video and several images.  
  Generate a \emph{descriptive caption} for the video that prominently features the components shown in the images.  
  Wrap your final text in \texttt{<caption>}\dots\texttt{</caption>} tags.  
  The caption must highlight the significance and role of these components throughout the video,  
  while omitting filler such as  
  “The scene unfolds with a whimsical and heartwarming narrative, emphasizing the simple joys of life through the Teddy Bear's endearing actions”.

  \vspace{10pt}
  %-----------------------------------------------------
  \textbf{Examples of Descriptive Captions}
  \begin{enumerate}[leftmargin=*, label=\arabic*., itemsep=4pt]
    \item Film quality, professional quality, rich details. The video begins to show the surface of a pond, and the camera slowly zooms in to a close-up. The water surface begins to bubble, and then a blonde woman is seen coming out of the lotus pond soaked all over, showing the subtle changes in her facial expression.
    \item A professional male diver performs an elegant diving maneuver from a high platform. Full-body side view captures him wearing bright red swim trunks in an upside-down posture with arms fully extended and legs straight and pressed together. The camera pans downward as he dives into the water below.
  \end{enumerate}

  \end{tcolorbox}

  \caption{Prompt template used to generate captions for our training data.}

  \label{fig:training_caption_generation_prompt}
\end{figure*}

\begin{figure*}[t]
  \centering
  \begin{tcolorbox}[
      width=\textwidth,
      enhanced,
      % breakable,                % allow graceful page breaks
      colback=yellow!6!white,   % softer background
      colframe=yellow!60!black,
      colbacktitle=yellow!60!black,
      coltitle=white,
      title={\Large\bfseries Video-Prompt Enhancement Output},
      fonttitle=\bfseries,
      sharp corners,
      boxrule=0.9pt,
      left=4mm, right=4mm, top=2mm, bottom=2mm
  ]
  \small  % compact text for better fit

  %-----------------------------------------------------
  \textbf{Task Description}\\[4pt]
  You will be given a prompt and several images for video generation.  
  You task is to make the prompt richer in description so the model can understand better.  
  Enclose your caption within \texttt{<caption></caption>} tags.  
  The caption must emphasize the significance and role of these components (and some description of each component) throughout the video.  
  Your caption should exclude unnecessary information such as  
  “The scene unfolds with a whimsical and heartwarming narrative, emphasizing the simple joys of life through the Teddy Bear's endearing actions”.

  \vspace{10pt}
  %-----------------------------------------------------
  \textbf{Example of a Descriptive Caption}
  \begin{enumerate}[leftmargin=*, label=\arabic*., itemsep=4pt]
    \item Film quality, professional quality, rich details.  
          The video begins to show the surface of a pond, and the camera slowly zooms in to a close-up.  
          The water surface begins to bubble, and then a blonde woman is seen coming out of the lotus pond soaked all over,  
          showing the subtle changes in her facial expression.
    \item A professional male diver performs an elegant diving maneuver from a high platform.  
          Full-body side view captures him wearing bright red swim trunks in an upside-down posture  
          with arms fully extended and legs straight and pressed together.  
          The camera pans downward as he dives into the water below.
  \end{enumerate}

  \vspace{10pt}
  %-----------------------------------------------------
  \textbf{Prompt to Optimize}\\[4pt]
  \{Insert your test prompt to optimize here\}

  \end{tcolorbox}

  \caption{Prompt template for test prompt enhancement.}

  \label{fig:test_video_prompt_enhanced}
\end{figure*}

\section{Details about User Study}
To ensure a smooth user study and annotation experience, we developed an HTML-based interface for participants to annotate and submit data. In this section, we describe the hiring platform, the job posting, and the design of the annotation interface.

\subsection{User Study Platform}
We recruit participants through Prolific,\footnote{\url{https://www.prolific.com/}}
 a research platform designed for user studies. Prolific offers an AI user study beta program that targets participants with experience in generative AI annotation.

To ensure quality, we apply screening filters to select participants with prior video annotation experience and fluent English proficiency, as understanding nuanced textual prompts is crucial for this task.

We hire 40 participants, each tasked with annotating five video sets, where each set contains generated outputs from four different models. The annotation process takes approximately 15 minutes per participant. Each is compensated \$5.50, reflecting the expected time and effort.

Our recruitment post and task instructions are shown in Figure~\ref{fig:annotation_instructions}.

\begin{figure*}[t]
\centering
\begin{tcolorbox}[width=\textwidth,
                  colback=yellow!10!white,
                  colframe=yellow!50!black,
                  title={\large\bfseries  Annotation Task Instructions},
                  fonttitle=\bfseries,
                  sharp corners,
                  boxrule=0.8pt,
                  left=3mm, right=3mm, top=2mm, bottom=2mm,
                  enhanced]  % ← borders + break lines nicely

\textbf{You will be presented with \underline{five} sets of short, AI-generated videos (\emph{5 s, no audio}).}

\medskip
Each set contains:
\begin{itemize}
  \item \textbf{Prompt} – textual description of the intended video (scene, objects, motion).
  \item \textbf{Reference Image} – split into two halves:
        \begin{itemize}
           \item \emph{Left side}: foreground objects that should appear in the video.
           \item \emph{Right side}: background scene to be integrated with the objects.
        \end{itemize}
  \item \textbf{Generated Videos (4 total)} – four model outputs attempting to fuse the objects with the background.
\end{itemize}

\bigskip
{\large\textbf{Your Task for Each Set}}

\medskip
\textbf{Step 1: Overall Ranking}
\begin{itemize}
  \item Watch \emph{all four} videos carefully.
  \item Rank them from best to worst based on overall quality and faithfulness to the prompt.
  \item Assign unique ranks (1 = best, 4 = worst).
\end{itemize}

\textbf{Step 2: Aspect Ratings}\par
After ranking, rate each video on a 1–5 scale (1 = very poor, 5 = excellent):
\begin{itemize}
  % \item \textbf{Object Number} – How many foreground objects are clearly visible?
  \item \textbf{Object Identity} – How well do objects retain their identity?
  \item \textbf{Scene / Background Identity} – How well is the background preserved?
  \item \textbf{Video Quality} – Overall realism and temporal coherence.
\end{itemize}

\bigskip
{\large\textbf{Notes}}
\begin{itemize}
  \item Evaluate \textbf{all four videos} in every set \emph{before} submitting answers.
  \item There are five sets in total (20 videos).
\end{itemize}
\end{tcolorbox}

\caption{Recruitment post for our user study.}
\label{fig:annotation_instructions}
\end{figure*}

\subsection{User Interface Details}

Participants first arrive at a login screen, where they enter their unique Prolific ID to match their responses with task-completion records. After authentication, they are presented with the textual prompt used to generate the videos, along with a composite reference image showing the required foreground objects and background. Below, four candidate videos are displayed in a randomized order to eliminate position bias. Participants then rank the videos based on overall quality, as shown in Figure~\ref{fig:interface1}.

Next, participants scroll down to rate each video on three criteria, Object Identity, Scene Identity, and Overall Quality, using a 5-point Likert scale (Figure~\ref{fig:interface2}).

\begin{figure*}[t]
  \centering
  \begin{subfigure}[t]{0.48\textwidth}
    \centering
    \includegraphics[width=\linewidth]{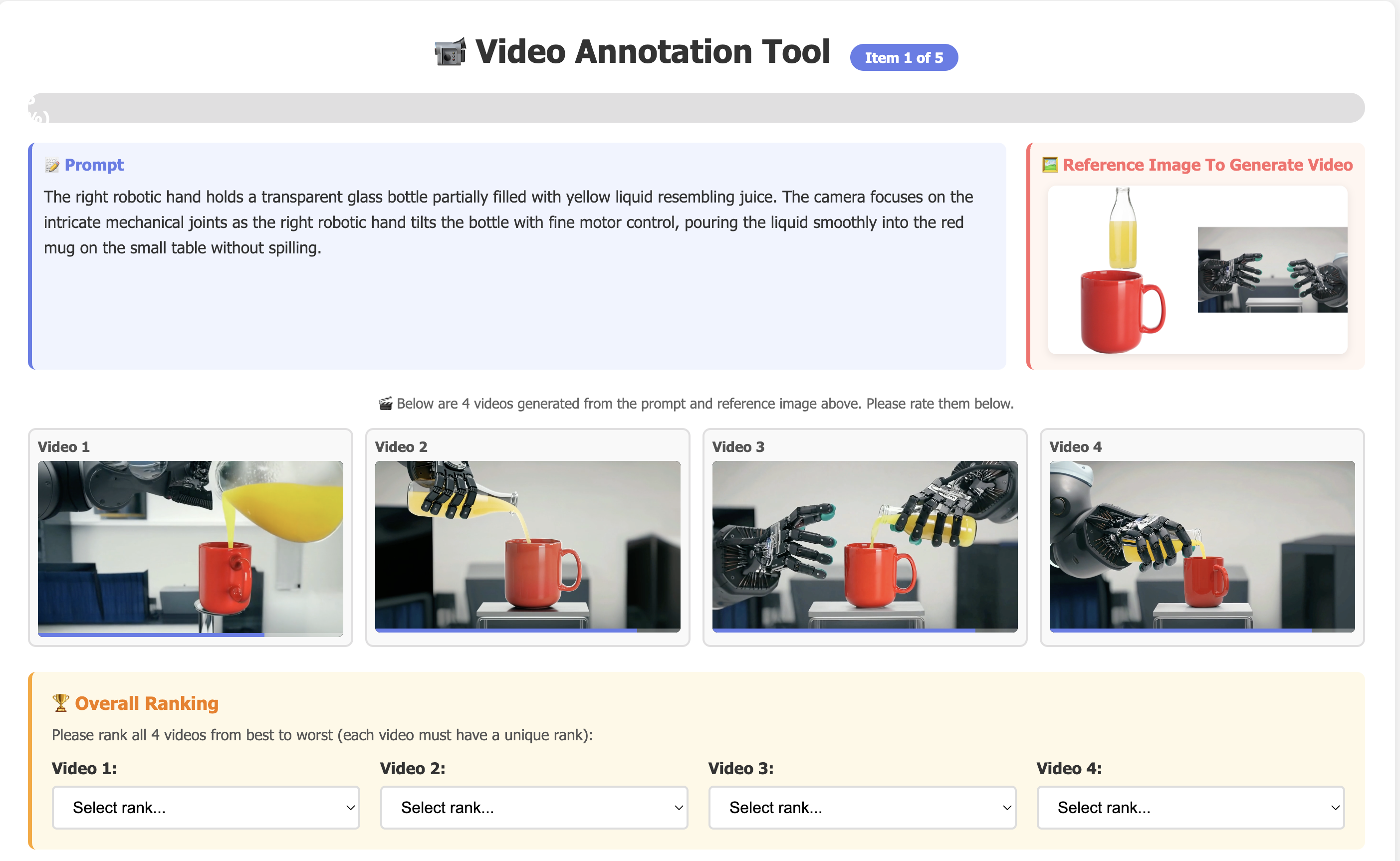}
    \caption{Users rank the overall quality of the four candidate videos.}
    \label{fig:interface1}
  \end{subfigure}\hfill
  \begin{subfigure}[t]{0.48\textwidth}
    \centering
    \includegraphics[width=\linewidth]{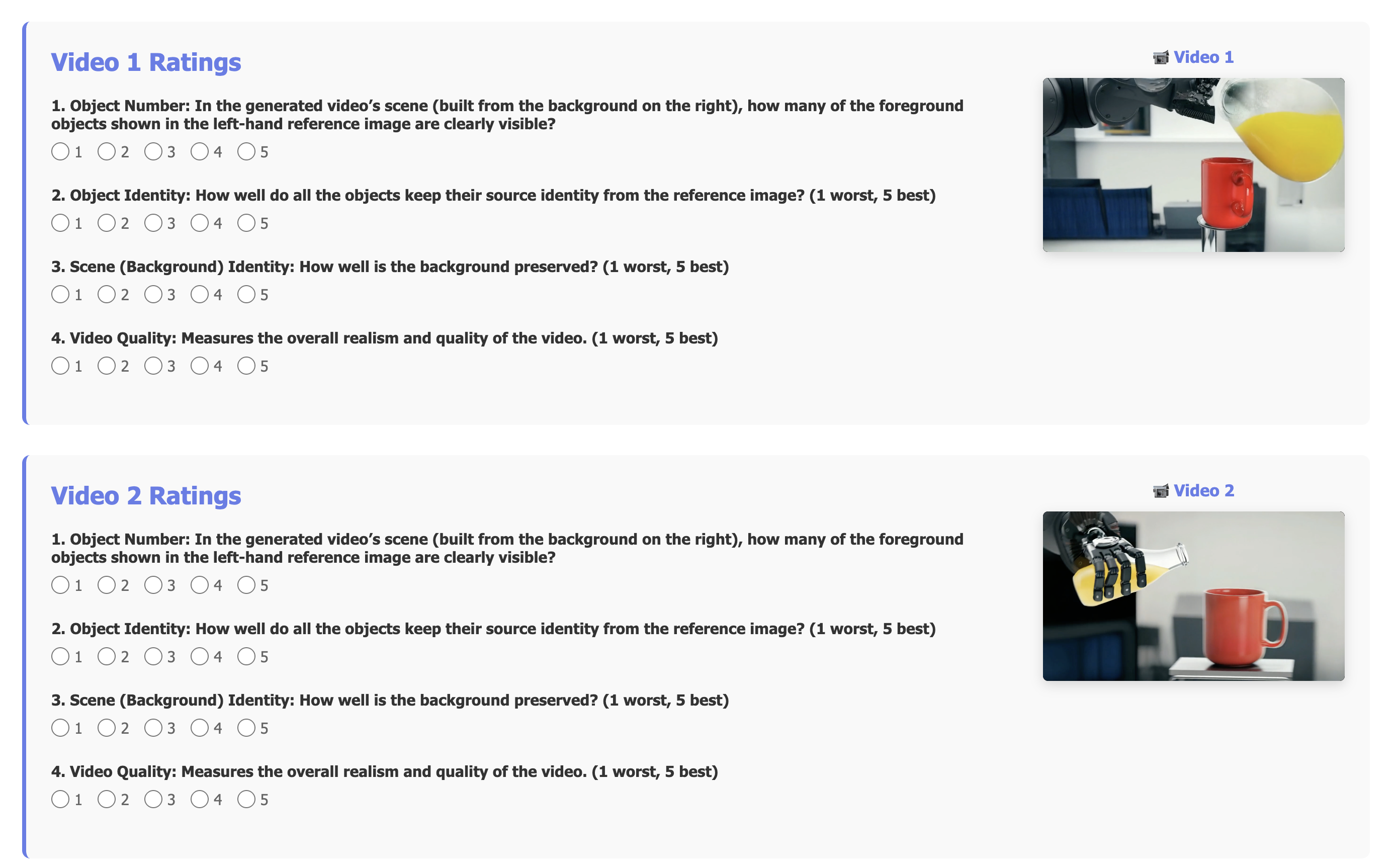}
    \caption{Users rate three specific aspects with a Likert scale 1-5.}
    \label{fig:interface2}
  \end{subfigure}
  \caption{Web-based annotation interface used in our user study. Part (a) collects a global quality ranking, while part (b) gathers detailed aspect-wise ratings for each video.}
  \label{fig:interface_overview}
\end{figure*}

\section{More Training and Inference Details}

We train LoRA modules of rank 128 for both high- and low-noise regime transformers in the base model Wan2.2-I2V-A14B. Training videos are resized to a resolution of 1344 $\times$ 768 with 81 frames. We use a batch size of 4 and optimize with AdamW \cite{loshchilov2017fixing}, setting the learning rate to $1 \times 10^{-4}$, $\epsilon = 1 \times 10^{-10}$, and a weight decay of $3 \times 10^{-2}$.

During inference, videos are generated at a resolution of 1280 $\times$ 720 with 81 frames, following the standard output format of Wan2.2-I2V-A14B based models.

\section{Generalization to First-Frame Layouts}
Although training uses a fixed first-frame layout (a), our model can generalize to unseen layouts in some cases. As shown in Fig.~\ref{fig:layout}, we evaluate three novel layouts (b), (c), and (d), beyond the training layout of cut-outs on the left and background on the right. The results suggest that our model interprets the first frame contextually rather than relying solely on the seen training layout.

\begin{figure}[t!]
    \centering
    
    \includegraphics[width=1.0\linewidth]{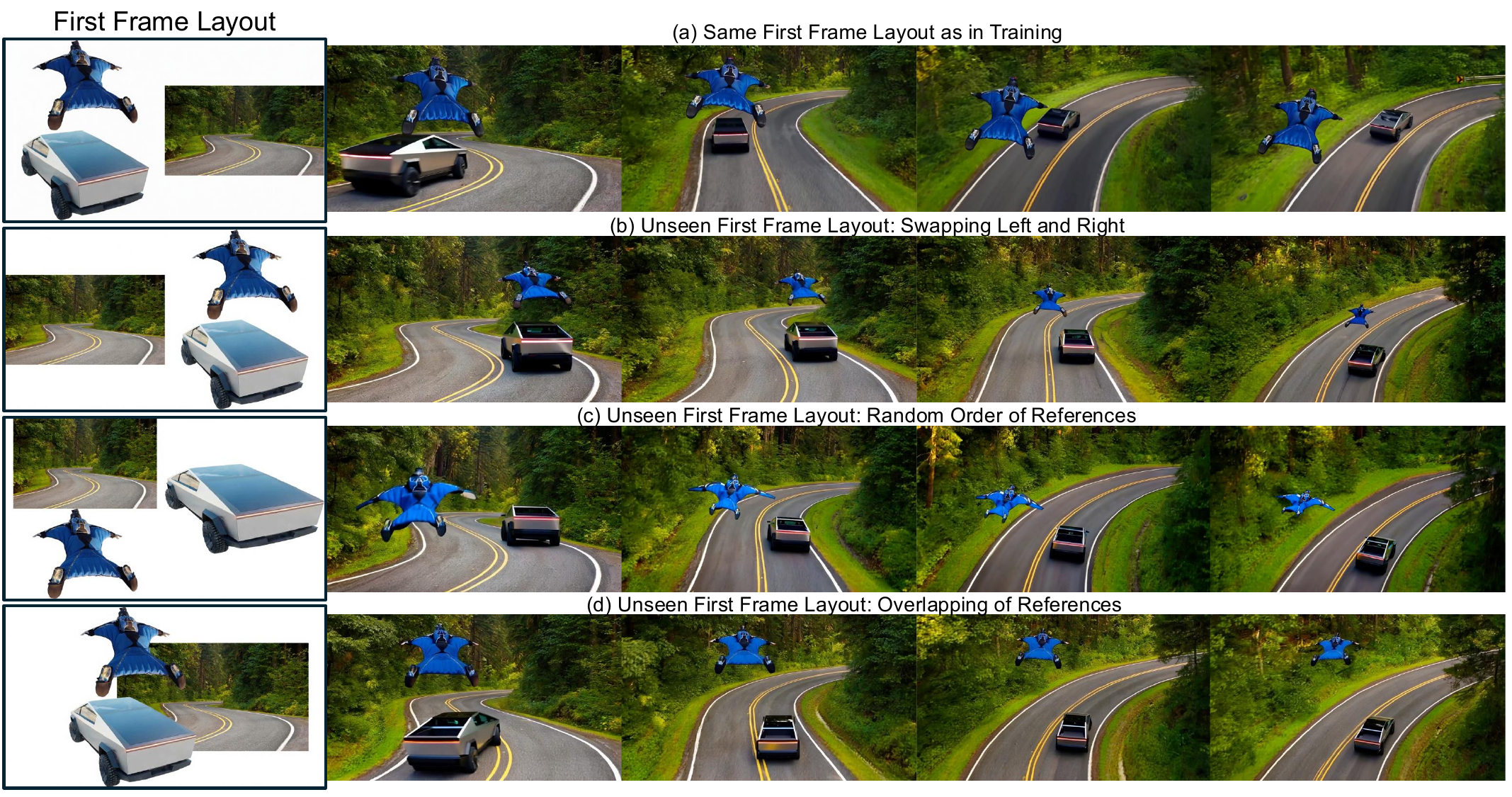} 
    % \vspace{-16pt}

    \small
    \vspace{-10pt}
    \caption{Generalization to spatial layouts in the first frame.}

    \label{fig:layout}
    \vspace{-13pt}
\end{figure}

\section{Visual Consistency Across Generated Videos from Different Reference Sources}
For all test results presented in the paper, the reference inputs are drawn from different sources rather than a single video. Visual consistency is maintained in the generated videos due to the pre-trained models' learned priors. For instance, in Fig.~\ref{fig:diffsource}, fine-grained shadows cast by a hand and bottle are correctly rendered on the teddy bear (shown by arrows).

\begin{figure}[t!]
    \centering
    
    \includegraphics[width=1.0\linewidth]{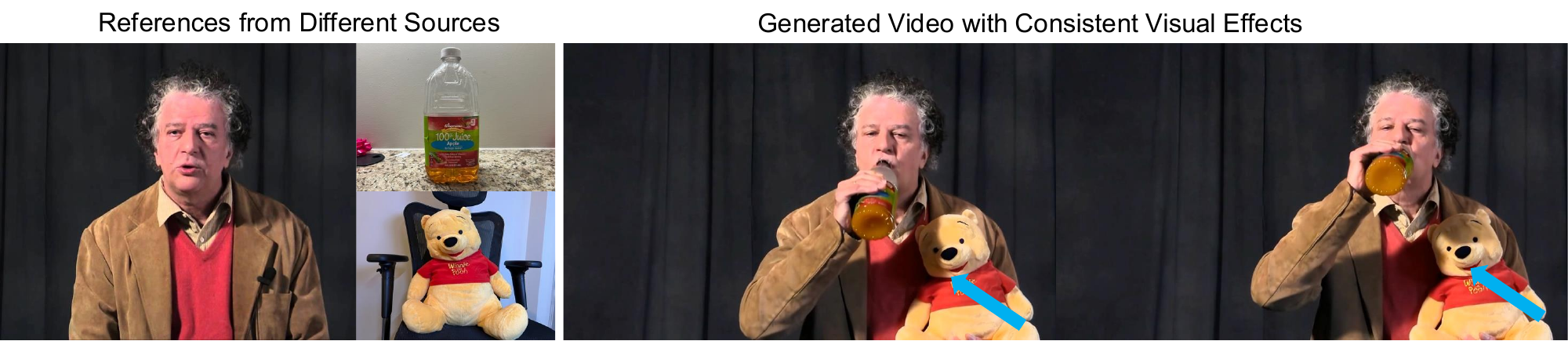} 
    % \vspace{-16pt}

    \small
    \vspace{-10pt}
    \caption{References from different sources.}

    \label{fig:diffsource}
    \vspace{-11pt}
\end{figure}

\section{Automatic Quantitative Metrics}
In Table \ref{tab:automatic}, we show standard automatic concept and VBench quantitative metrics: CLIP-I, DINO-I, CLIP-cap (text alignment), Motion Smoothness, Dynamic Degree (motion intensity), Aesthetic Quality, and Imaging Quality, to compare with baselines. These standard metrics further validate the effectiveness of our method.

\begin{table}[t!]
\centering

\resizebox{1.0\linewidth}{!}{%
\begin{tabular}{lccccccc}
\toprule
Model & CLIP-I$\uparrow$ & DINO-I$\uparrow$ & CLIP-cap$\uparrow$ & Motion Smoothness $\uparrow$ & Dynamic Degree$\uparrow$ & Aesthetic Quality$\uparrow$ & Imaging Quality$\uparrow$ \\
\midrule
Wan2.2-I2V-A14B & 0.66 & 0.42 & 33.2 & 0.96 & 9.75 & 0.82 & 0.61 \\
VACE & \textbf{0.68} & \textbf{0.46} & 33.6 & 0.97 & 7.78 & \textbf{0.92} & 0.65 \\
SkyReels-A2 & 0.66 & 0.43 & 33.1 & 0.96 & 12.93 & 0.91 & 0.72 \\
Ours & 0.67 & \textbf{0.46} & \textbf{34.00} & \textbf{0.98} & \textbf{14.64} & 0.85 & \textbf{0.73} \\
\bottomrule
\end{tabular}%
}
\vspace{-10pt}
\caption{Concept and VBench Automatic Quantitative Metrics}
\vspace{-20pt}

\label{tab:automatic}
\end{table}

\section{Explanation of the Transition Phrase}
The transition phrase $<$transition$>$ (e.g., “ad23r2 the camera view suddenly changes”) serves as a unique trigger in the text prompt. Paired with LoRA training, it enables the base model to learn to invoke latent abilities for scene cuts and reference fusion when encountered during inference. The choice of trigger can be arbitrary, as long as it is unique. This design is inspired by the use of unique trigger phrases in DreamBooth \cite{ruiz2023dreambooth}, but serves a fundamentally different purpose.

% % \clearpage 

% {
%     \small
%     \bibliographystyle{ieeenat_fullname}
%     \bibliography{main}
% }

\end{document}